%% file: main.tex
\documentclass[a4paper,twocolumn,10pt]{article}

\usepackage{glossaries}
\usepackage[utf8]{inputenc}
\usepackage[T1]{fontenc}
\usepackage[hidelinks]{hyperref}
\usepackage{url}
\usepackage{booktabs}
\usepackage{amsfonts}
\usepackage{nicefrac}
\usepackage{microtype}
\usepackage{geometry}
\usepackage{tikz}
\usepackage{pgfplotstable}
\usepackage[labelfont={small,bf},textfont=small]{caption}
\usepackage{subcaption}
\usepackage{amsmath}
\usepackage{mathtools}
\usepackage[separate-uncertainty=true,binary-units=true]{siunitx}
\usepackage{multirow}
\usepackage{authblk}
\usepackage[noabbrev,capitalise]{cleveref}  
\usepackage{rotating}
\usepackage{dsfont}
\usepackage{xifthen}
\usepackage{tabularx}
\usepackage{threeparttable}
\usepackage[numbers,sort&compress,authoryear]{natbib}
\usepackage{dirtree}
\usepackage{textgreek}

\setcitestyle{authoryear,open={[},close={]}}

\renewcommand{\cite}[1]{\citep{#1}}
\newcommand{\PARstart}[2]{{#1}{#2}}
\newcommand*\diff{\mathop{}\!\mathrm{d}}

\DeclareMathOperator{\argmax}{argmax}

\def\TableFontSize{\scriptsize}

\DeclareSIUnit\year{yr}

\usetikzlibrary{arrows.meta,decorations.pathreplacing,fadings,shapes,arrows,snakes,calc,patterns,positioning,decorations.pathmorphing,fit,spy}

\definecolor{blue}{RGB}{0,91,130}
\definecolor{red}{RGB}{185,70,60}
\definecolor{green}{RGB}{125,150,110}
\definecolor{orange}{HTML}{D7AA50}
\definecolor{purple}{HTML}{7A68A6}

\title{The Heidelberg spiking datasets for the systematic evaluation of spiking neural networks}

\author[1]{Benjamin Cramer}
\author[1]{Yannik Stradmann}
\author[1]{Johannes Schemmel}
\affil[1]{Kirchhoff Institute for Physics, University of Heidelberg, 
Germany}
\author[2]{Friedemann Zenke}
\affil[2]{Friedrich Miescher Institute for Biomedical Research, Basel,
Switzerland}

\input{acronyms}

\begin{document}

\twocolumn[
	\begin{@twocolumnfalse}
		\maketitle
		\begin{abstract}
			\input{abstract}
		\end{abstract}
		\hspace{1cm}
	\end{@twocolumnfalse}
]


\input{introduction}

\input{methods}

\input{results}
\input{discussion}

\appendix
\crefalias{section}{appendix}
\crefalias{subsection}{appendix}
\input{appendix}

\input{acknowledgments}

\bibliography{bibliography_cramer,bibliography_zenke}

\end{document}

%% file: acronyms.tex
\newacronym{bp}{BP}{backpropagation}
\newacronym{ann}{ANN}{artificial neural network}
\newacronym{rnn}{RNN}{recurrent neural network}
\newacronym{rsnn}{RSNN}{recurrently connected spiking neural network}
\newacronym{snn}{SNN}{spiking neural network}
\newacronym{bptt}{BPTT}{backpropagation through time}
\newacronym{svm}{SVM}{support vector machine}
\newacronym{lstm}{LSTM}{long short-term memory}
\newacronym{lif}{LIF}{leaky integrate-and-fire}
\newacronym{bm}{BM}{basilar membrane}
\newacronym{bc}{BC}{bushy cell}
\newacronym{hc}{HC}{hair cell}
\newacronym{api}{API}{application programming interface}
\newacronym{fft}{FFT}{fast Fourier transformation}
\newacronym{rms}{RMS}{root mean square}
\newacronym{rbf}{RBF}{radial basis function}
\newacronym{dvs}{DVS}{dynamic vision sensor}
\newacronym{nmnist}{N-MNIST}{neuromorphic MNIST}
\newacronym{xxx}{HD}{Heidelberg Digits}
\newacronym{sxxx}{SHD}{Spiking Heidelberg Digits}
\newacronym{google}{SC}{Speech Commands}
\newacronym{sgoogle}{SSC}{Spiking Speech Commands}
\newacronym{hdf5}{HDF5}{Hierarchical Data Format 5}
\newacronym{stp}{STP}{shortm-term plasticity}
\newacronym{stdp}{STDP}{spike-timing dependent plasticity}
\newacronym{xor}{XOR}{exclusive-OR}
\newacronym{flac}{FLAC}{Free Lossless Audio Codec}
\newacronym{cnn}{CNN}{convolutional neural networks}
\newacronym{relu}{ReLU}{rectified linear unit}
\newacronym{snu}{SNU}{spiking neural unit}

%% file: abstract.tex
Spiking neural networks are the basis of versatile and power-efficient
information processing in the brain.
Although we currently lack a detailed understanding of how
these networks compute,
recently developed optimization techniques 
allow us to instantiate increasingly complex functional spiking neural networks
in-silico.
These methods hold the promise to build more efficient non-von-Neumann computing
hardware and will offer new vistas in the quest of unraveling brain circuit function.
To accelerate the development of such methods, objective ways
to compare their performance are indispensable.
Presently, however, there are no widely accepted means for comparing the
computational performance of spiking neural networks.
To address this issue, we introduce two spike-based classification
datasets, broadly applicable to benchmark
both software and neuromorphic hardware implementations of spiking neural
networks.
To accomplish this, we developed a general audio-to-spiking conversion procedure
inspired by neurophysiology.
Further, we applied this conversion to an existing and a novel speech dataset.
The latter is the free, high-fidelity, and word-level aligned Heidelberg digit
dataset that we created specifically for this study.
By training a range of conventional and spiking classifiers, we show that
leveraging spike timing information within these datasets is essential for good
classification accuracy.
These results serve as the first reference for future performance comparisons
of spiking neural networks.

%% file: introduction.tex
\section{\label{sec:introduction}Introduction}

\glsdisp{snn}{\PARstart{S}{piking} neural networks (SNNs)} are biology's solution for fast and versatile information processing.
From a computational point of view, \glspl{snn} have several desirable properties:
They process information in parallel, are noise-tolerant, and highly energy efficient~\cite{boahen2017}.
Precisely which computations are carried out in a given biological \gls{snn} depends in large part on its connectivity structure.
To instantiate such functional connectivity in-silico, 
a growing number of \glspl{snn} training algorithms have been developed~\cite{zenke_superspike:_2018, pfeiffer_deep_2018, tavanaei_deep_2018,
bellec2018, shrestha_slayer:_2018, wozniak_deep_2020, neftci_surrogate_2019} 
both for conventional computers and neuromorphic hardware~\cite{schemmel2010,friedmann2016,furber2013,davies2018,moradi2018,roy_towards_2019}.
However, this diversity of learning algorithms urgently calls for
principled means to compare them.
Unfortunately, widely accepted benchmark datasets for \glspl{snn} that would
permit such comparisons are scarce~\cite{davies2019,roy_towards_2019}.
Hence, in this article, we seek to fill this gap by introducing two new broadly applicable classification datasets for \glspl{snn}.

In the following, we provide a brief motivation for why benchmarks are crucial before reviewing existing tasks that have been used to assess \gls{snn} performance in the past. 
By analyzing the strengths and shortcomings of these tasks, we motivate our specific choices for the datasets we introduce in this article.
Finally, we establish the first set of baselines by testing a range of conventional and \gls{snn} classifiers on these datasets.

\subsection{\label{subsec:why_benchmarks}Why benchmarks?}

The ultimate goal of a benchmark is to provide a quantitative unbiased way of comparing different approaches and methods to the same problem.
While each modeler usually works with a set of \textit{private} benchmarks,
tailored to their specific problem of study, it is equally important
to have shared benchmarks, which ideally everybody agrees to use, to allow for
unbiased comparison and to foster constructive competition between approaches~\cite{davies2019,roy_towards_2019}.

The last decades of machine learning research would be hard to imagine without the ubiquitous MNIST dataset~\cite{lecun1998}, for instance.
To process MNIST using a \glspl{snn}, it has to be transformed into spikes.
This transformation step puts comparability at risk by leaving fundamental design decisions to the modeler.
Presently, the \gls{snn} network community has a shortage of established
benchmarks that avoid the conversion step by directly providing spike trains to
the end-user.
By impeding the quantitative comparison between methods, the lack of suitable benchmarks has the potential to slow down the progress of the \gls{snn} research community as a whole.

Since community benchmarks are essential, then why is there little agreement on which benchmark to use?  
There are several possible reasons, but the most likely ones are the following:
First, an existing benchmark may be \textit{unobtainable}.
For instance, it could be unpublished, behind a paywall, or too difficult to use.
Second, a published benchmark might be \textit{tailored} to a specific problem
and, therefore, not general enough to be of interest to other researchers.
Third, a benchmark may be \textit{saturated}, which means that it is already solved with high precision by an existing method.
Naturally, this precludes the characterization of improvements over these approaches.
Finally, a benchmark could require extensive preprocessing. 

The question, therefore, is: What would an ideal benchmark dataset for learning
in \glspl{snn} be?
While this question is difficult, if not impossible, to answer, it is probably fair to say that an ideal benchmark should be at least unsaturated, require minimal preprocessing, be sufficiently general, easy to obtain, and free to use.

\subsection{\label{subsec:previous_work}Previous work}

There are no unified approaches to measuring performance in \glspl{snn} which is partially due to the numerous different learning approaches and architectures. 
\Gls{snn} architectures can coarsely be categorized into steady-state rate-coding and temporal coding networks, although also hybrids between the two exist. 
In steady-state rate-coding, \glspl{snn} approximate conventional analog neural networks by using an effective firing rate code in which both input and output firing rates remain constant during the presentation of a single stimulus~\cite{zylberberg_sparse_2011, neftci_event-driven_2017, pfeiffer_deep_2018}.
Inputs to the network enter as Poisson distributed spike trains with rates are proportional to the current input level.
Similarly, network outputs are given as a firing rate or spike count of designated output units. Because of these input-output specifications, steady-state rate-coding networks can often be trained using network translation~\cite{pfeiffer_deep_2018} and they can be tested on standard machine learning datasets (e.g.\ MNIST~\cite{lecun1998}, CIFAR10~\cite{krizhevsky2009}, or SVHN~\cite{netzer2011}).

The capabilities of \glspl{snn}, however, go beyond such rate-coding networks.
In temporally coding networks, input and output activity varies during the processing of a single input example.
Within this coding scheme, outputs can be either individual spikes~\cite{gutig_spike_2014, bohte_error-backpropagation_2002, mostafa_supervised_2018, comsa_temporal_2019}, 
spike trains with predefined firing times~\cite{memmesheimer_learning_2014}, 
varying firing rates~\cite{gilra_predicting_2017, nicola_supervised_2017, thalmeier_learning_2016},
or continuously varying quantities derived from output spikes.
The latter are typically defined as linear combinations of low-pass filtered spike trains~\cite{eliasmith_neural_2004, deneve_efficient_2016, abbott_building_2016, nicola_supervised_2017, gilra_predicting_2017}.


One of the simplest temporal coding benchmarks is the temporal \gls{xor} task, which exists in different variations~\cite{bohte_error-backpropagation_2002,huh_gradient_2018, abbott_building_2016}.
A simple \gls{snn} without hidden layers cannot solve this problem, similar to the Perceptron's inability to solve the regular \gls{xor} task.
Hence, the temporal \gls{xor} is commonly used to demonstrate that a specific method supports hidden-layer learning. 
In the temporal \gls{xor} task, a neural network has to solve a Boolean \gls{xor} problem in which the logical \textit{off} and \textit{on} levels correspond to early and late spike times respectively.
While the temporal \gls{xor} does require a hidden layer to be solved correctly, its intrinsic low-dimensionality and the low number of input patterns render this benchmark saturated. 
Therefore, its possibilities for quantitative comparison between training methods are limited.

To assess learning in a more fine-grained way, several studies have focused on \glspl{snn}' abilities to generate precisely timed output spike trains in more general scenarios~\cite{ponulak_supervised_2009,pfister_optimal_2006, florian_chronotron:_2012, mohemmed_span:_2012, memmesheimer_learning_2014,gardner_supervised_2016}.
To that end, it is customary to use several Poisson input spike trains to generate a specific target spike train.
Apart from regular (e.g.\ \citet{gardner_supervised_2016}), also random output spike trains with increasing length and Poisson statistics have been considered~\cite{memmesheimer_learning_2014}.
Similarly, the Tempotron~\cite{gutig_tempotron:_2006} uses an interesting hybrid approach in which random temporally encoded spike input patterns are classified into binary categories corresponding to spiking versus quiescence of a designated output neuron.
In the associated benchmark, task performance is measured as the number of binary patterns that can be classified correctly.
While mapping random input spikes to output spikes allows a fine-grained
comparison between methods, the aforementioned tasks lack a non-random structure.

Finally, some datasets for n-way classification were born out of practical engineering needs. 
The majority of these datasets are based on the output of neuromorphic sensors like, for instance, the \gls{dvs}~\cite{lichtsteiner_128times128_2008} or the silicon cochlea~\cite{anumula_feature_2018}.
An early example of such a dataset is Neuromorphic MNIST~\cite{orchard_converting_2015}, which was generated by a \gls{dvs} recording MNIST digits that were projected on a screen.
The digits were moved at certain intervals to elicit spiking responses in the \gls{dvs}.
The task is to identify the corresponding digits from the elicited spikes.
This benchmark has been used widely in the \gls{snn} community.
However, being based on the MNIST dataset it is nearing saturation.
The DASDIGITS dataset~\cite{anumula_feature_2018} was created by processing the TIDIGITS with a 64~channel silicon cochlea.
Unfortunately, the license requirements for the derived dataset are not entirely clear as the TIDIGITS are released under a proprietary license.
Moreover, because TIDIGITS contains sequences of spoken digits, the task goes beyond a straight-forward n-way classification problem and therefore is beyond the scope for many current \gls{snn} implementations.
More recently, IBM has released the DVS128 Gesture Dataset~\cite{amir_low_2017} under a Creative Commons license.  
The dataset consists of numerous \gls{dvs} recordings of 11~unique hand gestures performed by different persons under varying lighting conditions.
The spikes in this dataset are provided as a continuous data stream, which makes extensive cutting and preprocessing necessary.
Finally, the 128$\times$128 pixel size renders this dataset computationally expensive unless additional preprocessing steps such as downsampling are applied.

In this article, we sought to generate two widely applicable \gls{snn} benchmarks with comparatively modest computational requirements.
Thus, we focused on audio signals of spoken words due to their natural temporal dimension and lower bandwidth compared to video data and developed a processing framework to convert these audio data into spikes.
Using this framework, we generated two spike-based datasets for speech classification and keyword spotting that are not saturated by current methods. 
Moreover, solving these problems with high accuracy requires taking into account spike timing.

%% file: methods.tex
\section{\label{sec:methods}Methods}

To improve the quantitative comparison between \glspl{snn}, we have created two
large spike-based classification datasets from audio data.
Specifically, we recorded the \glsfirst{xxx} dataset for this purpose and used the published \gls{google} dataset by the TensorFlow and AIY teams~\cite{warden2018}.
In the following, we describe the datasets (\cref{subsec:datasets}), the audio-to-spike conversion (\cref{subsec:preprocessing}), and the data format used for publication (\cref{subsec:dataformat}).
We close with a depiction of the \gls{snn} model (\cref{subsec:spiking_nets}).
The non-spiking classifiers are outlined in \cref{subsec:controls}.
All reported error measures in this work correspond to the standard deviation of 10 experiments.

\subsection{\label{subsec:datasets}Audio datasets}

In the following, we consider the \gls{xxx} (\cref{subsubsec:xxx}) and the \gls{google} dataset (\cref{subsubsec:speech_commands}).
While the \gls{xxx} were optimized for recording quality and precise audio alignment, the \gls{google} are intended to closely mimic real-world conditions for key-word spotting on mobile devices.

\subsubsection{\label{subsubsec:xxx}\Acrlong{xxx}}

\begin{figure}[t]
	\centering
	\input{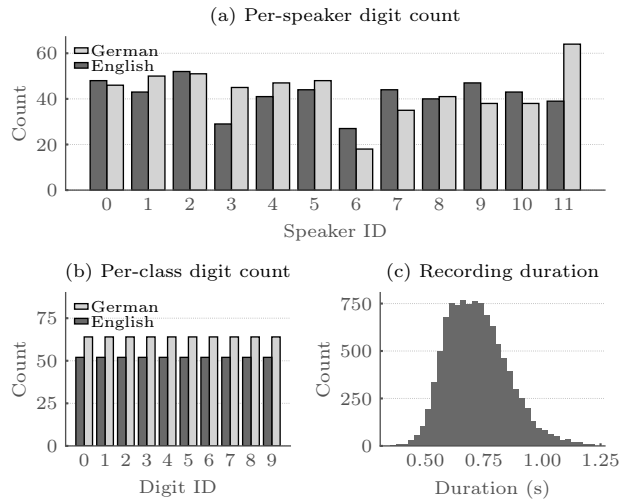}
	\caption{%
		\textbf{The \glsfirst{xxx} have a balanced class count and variable temporal duration.}
	    	The \gls{xxx} consist of 10420 recordings of spoken digits ranging from \textit{zero} to \textit{nine} in English and German language.
		(a)~Histogram of per-speaker digit counts. 
		Variable numbers of digits are available for each speaker and each language.
		(b)~Histogram of per-class digit counts.
		The dataset is balanced in terms of digits within each language.
		(c)~Histogram of audio recording durations.
		The \gls{xxx} audio recordings were cut for minimal duration to keep computation time at bay.
	}
	\label{fig:digits}
\end{figure}

The \gls{xxx}\footnote{\label{urlnote}\url{https://compneuro.net}} consist of approximately 10k~high-quality recordings of spoken digits ranging from \textit{zero} to \textit{nine} in English and German language.
In total \SI{12}{} speakers were included, six of which were female and six male.
The speaker ages ranged from \SI{21}{\year} to \SI{56}{\year} with a mean of \SI{29(9)}{\year}.
We recorded around \SI{40}{} digit sequences for each language with a total digit count of \SI{10420}{} (cf. \cref{fig:digits}).

The digits were acquired in sequences of ten successive digits.
Recordings were performed in a sound-shielded room at the Heidelberg University Hospital with three microphones; two AudioTechnica Pro37 in different positions and a Beyerdynamic M201 TG (\cref{fig:overview}).
Digitized by a Steinberg MR816 CSX audio interface, recordings were made in WAVE format with a sample rate of \SI{48}{\kilo\hertz} and \SI{24}{\bit} precision.

To improve the yield of the following automated processing, a manual pre-selection and cutting of the raw audio tracks were performed accompanied by conversion to \gls{flac} format.
The cleaned-up tracks were externally mastered~\cite{devcore2019}.
The cutting times of the digit sequences were determined using a gate with speaker-dependent threshold and release time which were optimized by the blackbox-optimizer described in~\citet{knysh2016}.
The loss function was designed to produce 10 single files with the lowest possible threshold and shortest gate opening to prevent unnecessary computation during successive analysis and modeling.
Additionally, speaker-specific ramp-in and ramp-out times were determined by visual inspection.
The final digit files differ in duration due to speaker differences (\cref{fig:digits}).
\SI{30}{\milli\second} Hann windows were applied to the start and end of the peak normalized audio signals as further processing stages involve the computation of \glspl{fft}.


To separate the data into training and test sets, we held out two speakers exclusively for the test set.
The remainder of the test set was filled with samples (\SI{5}{\percent} of the trials) from speakers also present in the training set.
This division allows to assess a trained network's ability to generalize across speakers.

\subsubsection{\label{subsubsec:speech_commands}\Acrlong{google}}

The \gls{google}\footnote{\url{https://www.tensorflow.org}} are composed of \SI{1}{\second} WAVE-files with \SI{16}{\kilo\hertz} sample rate containing a single English word each~\cite{warden2018}.
It is published under \textit{Creative Commons BY 4.0} license and contains words spoken by \SI{1864}{} speakers.
In this study, we considered version 0.02 with \SI{105829}{} audio files, in which a total of 24 single word commands (\textit{Yes}, \textit{No}, \textit{Up}, \textit{Down}, \textit{Left}, \textit{Right}, \textit{On}, \textit{Off}, \textit{Stop}, \textit{Go}, \textit{Backward}, \textit{Forward}, \textit{Follow}, \textit{Learn}, \textit{Zero}, \textit{One}, \textit{Two}, \textit{Three}, \textit{Four}, \textit{Five}, \textit{Six}, \textit{Seven}, \textit{Eight}, \textit{Nine}) were repeated about five times per speaker, whereas ten auxiliary words (\textit{Bed}, \textit{Bird}, \textit{Cat}, \textit{Dog}, \textit{Happy}, \textit{House}, \textit{Marvin}, \textit{Sheila}, \textit{Tree}, and \textit{Wow}) were only repeated approximately once.
Partitioning into training, testing and validation dataset was done by a hashing
function as described in~\citet{warden2018}.

For all our purposes, we applied a \SI{30}{\milli\second} Hann window to the start and end of each waveform.
Most importantly, throughout this article we consider top one classification
performance on all 35~different classes which
is more difficult than the originally proposed key-word spotting task on  only a
subset of 12~classes (10~key-words, unknown word, and silence).
However, the data can still be used in the originally intended keyword spotting way.

\subsection{\label{subsec:preprocessing}Spike conversion}

\glsreset{bm}
\glsreset{hc}
\glsreset{bc}

\begin{figure}[tbp]
	\centering
	\scalebox{0.52}{\input{graphics/overview}}
	\caption{%
		\textbf{Processing pipeline for the \gls{xxx} and the \gls{google} dataset.}
		(a)~The \gls{xxx} are recorded in a sound-shielded room.
		(b)~Afterwards, the resulting audio files are cut and mastered.
		(c)~The \gls{xxx} as well as the \gls{google} are fed through a
		hydrodynamic \acrlong{bm} model.
		(d)~\Acrlong{bm} decompositions are converted to phase-coded spikes by
		use of a transmitter-pool based \acrlong{hc} model.
		(e)~The phase-locking is increased by combining multiple spiketrains of
		\acrlongpl{hc} at the same position of the \acrlong{bm} in a single \acrlong{bc}.
	}
	\label{fig:overview}
\end{figure}

The audio files described above served as the basis for our spiking datasets.
Audio data were converted into spikes using an artificial model\footnote{\url{https://github.com/electronicvisions/lauscher}} of the inner ear
and parts of the ascending auditory pathway (\cref{fig:overview}; \cref{sec:inner-ear}). 
This biologically inspired model effectively performs similar signal processing steps
as customary spoken language processing applications~\cite{huang_spoken_2001}.
First, a hydrodynamic \gls{bm} model (\cref{subsubsec:bm}) causes spatial frequency dispersion, which is comparable to computing a spectrogram with Mel-spaced filter banks.
Second, these separated frequencies are converted to firing rates through a
biologically motivated transmitter pool based \gls{hc} model
(\cref{subsubsec:hc}), which adds refractory effects, and a layer of
\glspl{bc} (\cref{subsubsec:bc}) that increase phase locking
(cf.\ \cref{fig:overview}).
All model parameters were chosen to mimic biological findings, thereby reducing
the amount of free parameters (see~\cref{tab:params}).

Overall, the inner-ear model approximates the spiking activity observed in the auditory system while retaining a low computational cost.
This biologically inspired conversion allowed us to sidestep the issue of user-specific audio-to-spike transformation, which can confound comparability, and served as the basis for our benchmark datasets.

\subsection{\label{subsec:dataformat}Event-based data format}

We used an event-based representation of spikes in the \gls{hdf5} to facilitate the use of the datasets and to simplify the access to a broader community, .
This choice was to ensure short download times and ease of access from most common programming environments.
For each partition and dataset, we provide a single \gls{hdf5} file which holds spikes, digit labels, and additional meta information.
We made these files publicly available\footnotemark[2]~\cite{cramer2019}, together with supplementary information on the general usage as well as code snippets.
A single file is organized as follows:
\begin{footnotesize}
	\dirtree{%
		.1 root.
		.2 spikes.
		.3 times[][]\DTcomment{VLArray of Arrays holding spike times}.
		.3 units[][]\DTcomment{VLArray of Arrays holding spike units}.
		.2 labels[]\DTcomment{Array of digit IDs}.
		.2 extra.
		.3 speaker[]\DTcomment{Array of speaker IDs}.
		.3 keys[]\DTcomment{Array of digit description strings}.
		.3 meta\_info.
		.4 gender[]\DTcomment{Array of speaker genders}.
		.4 age[]\DTcomment{Array of speaker ages}.
		.4 body\_height[]\DTcomment{Array of speaker body heights}.
	} 
\end{footnotesize}
\vspace{0.2cm}

In more detail, each element $i$ in \textit{keys} describes the transformation between the digit ID $i$ and the spoken words.
Further, the entry $i$ of each array in \textit{meta\_info} corresponds to the information for speaker $i$.
The \textit{meta\_info} is only available for \gls{sxxx}.

\subsection{\label{subsec:spiking_nets}Spiking network models}

We trained networks of \gls{lif} neurons with surrogate gradients and \gls{bptt} using supervised loss functionsTo establish a performance reference on the two spiking datasets.
In the following we give a description of the network architectures (\cref{subsubsec:network}), followed by the applied neuron and synapse model (\cref{subsubsec:neuron}).
We close with a depiction of the weight initialization (\cref{subsubsec:init}), the supervised learning algorithm (\cref{subsubsec:supervised}) and the loss function (\cref{subsubsec:loss}) as well as the regularization techniques (\cref{subsubsec:regularization}).

\subsubsection{\label{subsubsec:network}Network model}

The spiketrains emitted by the $N_\mathrm{ch} = \SI{700}{}$ \glspl{bc} were used to stimulate the actual classification network.
In this manuscript, we trained both feed-forward and recurrent networks; each hidden layers containing $N=128$ \gls{lif} neurons.
For all network architectures, the last layer was accompanied by a linear readout consisting of leaky integrators which did not spike.

\subsubsection{\label{subsubsec:neuron}Neuron and synapse models}

We considered \gls{lif} neurons where the membrane potential $u_i^{(l)}$ of the $i$-th neuron in layer $l$ obeys the differential equation:
\begin{equation}
	\tau_\mathrm{mem} \frac{\mathrm{d}u^{(l)}_i}{\mathrm{d}t} = - [u_i^{(l)}(t) - u_\mathrm{leak}] + RI_i^{(l)}(t) \label{eq:membrane} \, ,
\end{equation}
with the membrane time constant $\tau_\mathrm{mem}$, the input resistance $R$, the leak potential $u_\mathrm{leak}$, and the input current $I^{(l)}_i(t)$.
Spikes were described by their firing time.
The $k$-th firing time of neuron $i$ in layer $l$ is denoted by $\prescript{}{k}{t}_i^{(l)}$ and defined by a threshold criterion:
\begin{equation}
	\prescript{}{k}{t}_i^{(l)} : u_i^{(l)}(\prescript{}{k}{t}_i^{(l)}) \geq u_\mathrm{thres}.
\end{equation}
Immediately after $\prescript{}{k}{t}_i^{(l)}$, the membrane potential is set to the leak potential $u_i^{(l)}(t) = u_\mathrm{leak}$.
The synaptic input current onto the $i$-th neuron in layer $l$ was generated by the arrival of presynaptic spikes from neuron $j$, $S_j^{(l)}(t) = \sum_{k} \delta(t - \prescript{}{k}{t}_j^{(l)})$.
A common first-order approximation to model the time course of synaptic currents are exponentially decaying currents which sum linearly~\cite{gerstner2002}:
\begin{align}
	\frac{\mathrm{d}I_i^{(l)}}{\mathrm{d}t} = &- \frac{I_i^{(l)}(t)}{\tau_\mathrm{syn}} \nonumber \\
						  &+ \sum_j W_{ij}^{(l)} S_j^{(l-1)}(t)
						   + \sum_j V_{ij}^{(l)} S_j^{(l)}(t) \label{eq:synapse} \, ,
\end{align}
where the sum runs over all presynaptic partners $j$ and $W_{ij}^{(l)}$ are the corresponding afferent weights from the layer below.
The $V_{ij}^{(l)}$ resemble the recurrent connections within each layer.

In this work, the reset was incorporated in \cref{eq:membrane} through an extra term:
\begin{align}
	\frac{\mathrm{d}u^{(l)}_i}{\mathrm{d}t} = &- \frac{u_i^{(l)} + u_\mathrm{leak} - RI_i^{(l)}}{\tau_\mathrm{mem}} \nonumber \\
						  &+ S_i^{(l)}(t) (u_\mathrm{leak} - u_\mathrm{thres}) \, .
\end{align}
To formulate the above equations in discrete time for time step $n$ and stepsize
$\delta t$ over a duration $T = n\cdot \delta t$, the output spiketrain
$S_i^{(l)}[n]$ of neuron $i$ in layer $l$ at time step $n$ is expressed as a
nonlinear function of the membrane potential $S_i^{(l)}[n] = \Theta\left(u_i^{(l)}
- u_\mathrm{thres}\right)$ with the Heavyside function $\Theta$.
For small time steps $\delta t$ we can express the synaptic
current in discrete time as follows:
\begin{align}
	I_i^{(l)}[n + 1] &= \kappa I_i^{(l)}[n] \nonumber \\
			 &+ \sum_j W_{ij}^{(l)}S_j^{(l)}[n] + \sum_j
			 V_{ij}^{(l)}S_j^{(l)}[n] \, . \label{eq:input_dis} 
\end{align}
Further, by asserting $u_\mathrm{leak} = 0$ and $u_\mathrm{thres} = 1$, the
membrane potential can be written compactly as:
\begin{align}
	u_i^{(l)}[n + 1] &= \lambda u_i^{(l)}[n] (1-S_i^{(l)}[n]) + (1-\lambda)
	I_i^{(l)}[n] \, , \label{eq:membrane_dis} 
\end{align}
where we have set $R = (1-\lambda)$ and introduced the constants $\kappa \equiv \exp{(-\delta t /
\tau_\mathrm{syn})}$ and $\lambda \equiv \exp{(-\delta t / \tau_\mathrm{mem})}$.

\subsubsection{\label{subsubsec:init}Weight initialization}
In all our spiking network simulations we use Kaiming's uniform initialization~\cite{he_delving_2015} for the weights $W_{ij}$ and $V_{ij}$.
Specifically, the initial weights were drawn independently from a uniform distribution
$\mathcal{U}(-\sqrt{k}, \sqrt{k})$ with $k = {(\text{\# afferent connections})}^{-1}$.

\subsubsection{\label{subsubsec:supervised}Supervised learning}

The goal of learning was to minimize a cost function $\mathcal{L}$ over the entire dataset.
To achieve this, surrogate gradient descent was applied which modifies the network parameters $W_{ij}$:
\begin{equation}
	W_{ij} \leftarrow W_{ij} - \eta \frac{\partial \mathcal{L}}{\partial W_{ij}} \, , \label{eq:surrogate_gradient}
\end{equation}
with the learning rate $\eta$.
In more detail, we used custom PyTorch~\cite{paszke2017} code implementing the
\glspl{snn}. 
Surrogate gradients were computed using PyTorch's automatic
differentiation capabilities by overloading
the derivative of the spiking nonlinearity with a differentiable
function as described previously~\cite{neftci_surrogate_2019,
zenke_remarkable_2020}.
An instructive example of such an implementation in PyTorch can be found 
online \footnote{\url{https://github.com/fzenke/spytorch}}.
Specifically, we chose a fast sigmoid to calculate the surrogate gradient:
\begin{equation}
	\sigma(u_i^{(l)}) = \frac{u_i^{(l)}}{1 + \beta \vert u_i^{(l)} \vert} \, , \label{eq:fast_sigmoid}
\end{equation}
with the steepness parameter $\beta$.

\subsubsection{\label{subsubsec:loss}Loss functions}

We applied a cross entropy loss to the activity of the readout layer $l=L$.
On data with $N_\mathrm{batch}$ samples and $N_\mathrm{class}$ classes, $\left\{(\mathbf{x}_s, y_s)\vert s=1,...,N_\mathrm{batch}; y_s\in\{1, ..., N_\mathrm{class}\}\right\}$ it takes the form:
\begin{multline}
	\mathcal{L} = -\frac{1}{N_\mathrm{batch}}\sum_{s=1}^{N_\mathrm{batch}} \mathds{1}(i = y_s) \\
	\cdot\log\left\{\frac{\exp{\left(u_i^{(L)}[\tilde{n}_i]\right)}}{\sum_{i=1}^{N_\mathrm{class}} \exp{\left(u_i^{(L)}[\tilde{n}_i]\right)}}\right\} \, , \label{eq:cross_entropy}
\end{multline}
with the indicator function $\mathds{1}$.
We tested to the following two choices for the time step $\tilde{n}$ (\cref{fig:loss_schematic}):
For the \textit{max-over-time} loss, the time step with maximal membrane potential for each readout unit was considered $\tilde{n}_i = \argmax_n u_i^{(L)}[n]$.
In contrast, the last time step $T$ for all samples was chosen for each readout neuron $\tilde{n}_i = T$ in case of the \textit{last-time-step} loss.
We minimized the cross entropy in \cref{eq:cross_entropy} using the Adamax optimizer~\cite{kingma2014}.

\subsubsection{\label{subsubsec:regularization}Regularization}

For our experiments, we added synaptic regularization terms to the loss function to avoid pathologically high or low firing rates.
In more detail, we used two different regularization terms:
As a first term, we used a per neuron lower threshold spike count regularization of the form:
\begin{multline}
	\mathcal{L}_1 = \frac{s_l}{N_\mathrm{batch} + N} \sum_{s=1}^{N_\mathrm{batch}} \sum_{i=1}^N \left[ \max \left\{0, \vphantom{\sum_{n=1}^T}\right. \right. \\
	\left. \left. \frac{1}{T} \sum_{n=1}^T S_i^{(l)}[n] - \theta_l \right\}\right]^2 \, ,
\end{multline}
with strength $s_l$, and threshold $\theta_l$.
Second, we used an upper threshold mean population spike count regularization:
\begin{multline}
	\mathcal{L}_2 = \frac{s_u}{N_\mathrm{batch}} \sum_{s=1}^{N_\mathrm{batch}} \left[ \max \left\{0, \vphantom{\sum_{n=1}^T}\right. \right. \\
	\left. \left. \frac{1}{N} \sum_{i=1}^N \sum_{n=1}^T S_i^{(l)}[n] - \theta_u \right\}\right]^2 \, ,
\end{multline}
with strength $s_u$, and threshold $\theta_u$.



%% file: graphics/overview.tex
\begin{tikzpicture}[spy using outlines={circle,black,magnification=4,size=2.2cm, connect spies}]
	\node (recording) {\includegraphics[width=0.15\textwidth, angle=90]{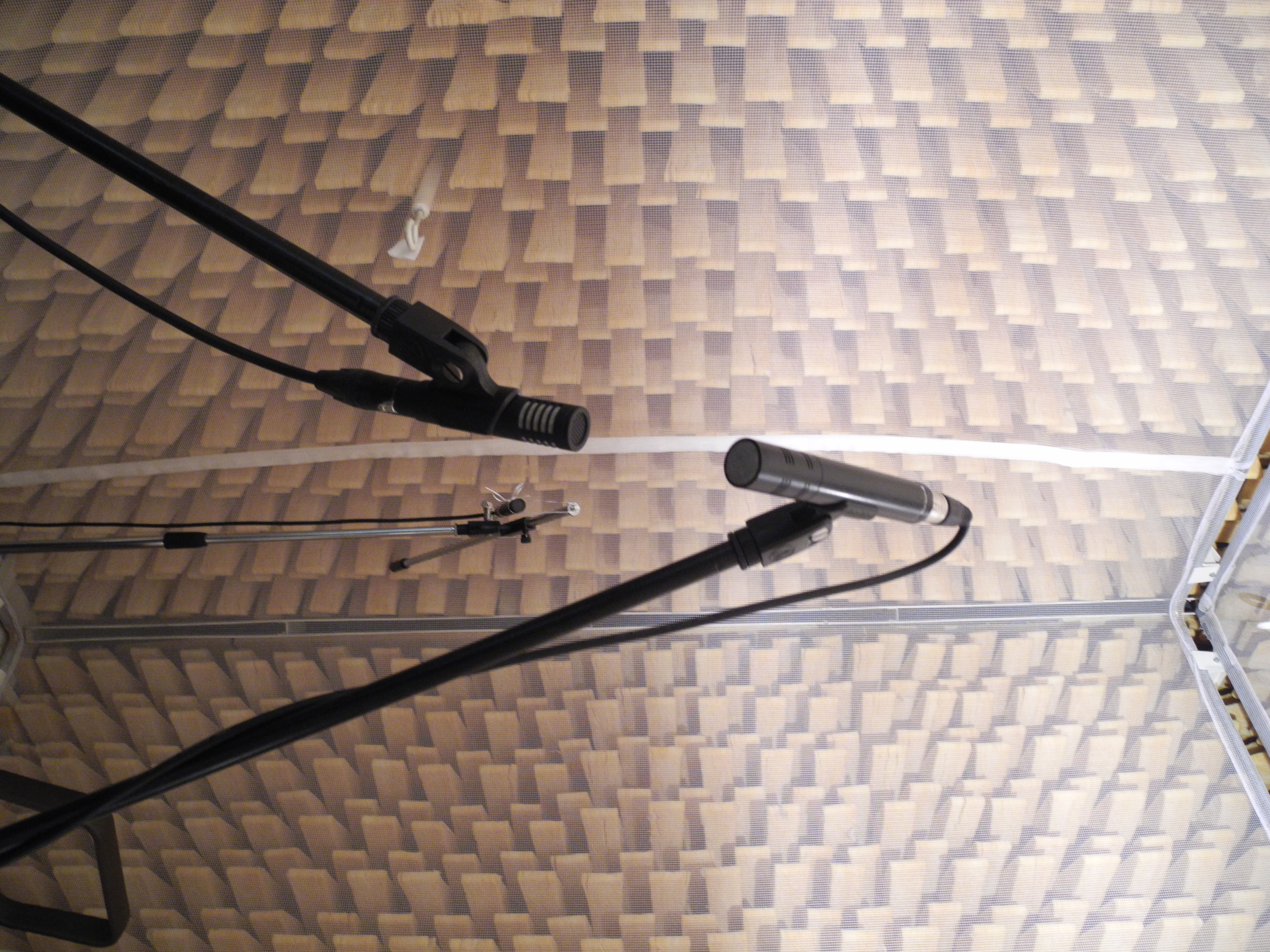}};
	\node (recording_caption) [above of=recording, yshift=0.7cm] {(a) Recording};
	\node (processing) [right of=recording, xshift=2.05cm] {\input{figures/raw.pgf}};
	\node (processing_caption) [above of=processing, yshift=0.7cm] {(b) Cutting};
	
	\node (bm) [right of=processing, xshift=2.7cm]{\includegraphics{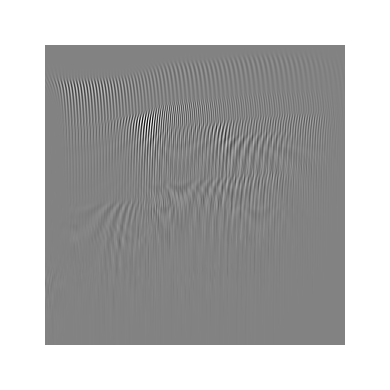}};
	\node (hc) [right of=bm, xshift=2.7cm] {\includegraphics{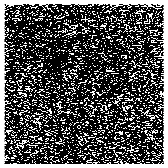}};
	\node (bc) [right of=hc, xshift=2.5cm] {\includegraphics{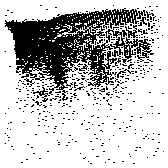}};
	
	\node (bm_caption) [above of=bm, yshift=0.7cm] {(c) \Acrlong{bm}};
	\node (hc_caption) [above of=hc, yshift=0.7cm] {(d) \Acrlongpl{hc}};
	\node (bc_caption) [above of=bc, yshift=0.7cm] {(e) \Acrlongpl{bc}};

	\node (google) [below of=processing, yshift=-3.0cm] {\input{figures/google.pgf}};

	\draw[very thick, -latex] (recording) -- (processing);
	\draw[very thick, -latex] (processing) -- (bm);
	\draw[very thick, -latex] (google) -- (bm);
	\draw[very thick, -latex] (bm) -- (hc);
	\draw[very thick, -latex] (hc) -- (bc);

	\node[draw,dotted,label=\Acrlong{xxx},fit=(recording) (processing_caption) (processing)] {};
	\node[draw,dotted,label=\Acrlong{google},fit=(google)] {};

	\spy on (6.5,0.5) in node [left] at (10,-3.5);
	\spy on (10.2,0.5) in node [left] at (14,-3.5);
\end{tikzpicture}

%% file: results.tex
\section{\label{sec:results}Results}

\begin{figure}[t]
	\centering
	\input{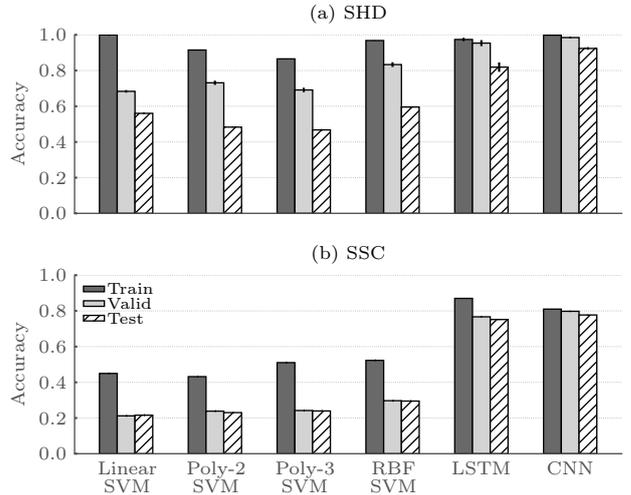}
	\caption{%
		\textbf{Temporal information is essential to classify the \acrshort{sxxx} and the \acrshort{sgoogle} datasets with high accuracy.}
		\textbf{(a)}~Bar graph of classification accuracy for different \acrshortpl{svm} trained on spike count vectors and \acrshort{lstm} as well as \acrshort{cnn} classifiers trained on the binned spiketrains of the \acrshort{sxxx} dataset.
		Classification accuracy on \acrshort{sxxx} is substantially higher for \acrshortpl{lstm} and \acrshortpl{cnn} which also show a lower degree of overfitting.
		\textbf{(b)}~Same as in (a), but showing performance on \acrshort{sgoogle}.
		\Acrshortpl{lstm} and \acrshortpl{cnn} with access to temporal information outperform the \acrshort{svm} classifiers by a large margin.
	}
	\label{fig:control}
\end{figure}

To analyze the relevance of our newly created spiking datasets  
we first sought to establish that the datasets were not saturated and that spike timing information is essential to solve the tasks with high accuracy.
To test this, we first generated a reduced version of the datasets in which we removed all temporal information.
To that end, we computed spike count patterns from both datasets, which, by design, do not contain temporal information about the stimuli.
Using these reduced spike count datasets, we then trained different linear and
nonlinear \gls{svm} classifiers (\cref{subsubsec:SVM}) and measured their classification performance on the respective test sets. 
We found that while a linear \gls{svm} readily overfitted the data in the case of \gls{sxxx}, its test performance only marginally exceeded the \SI{55}{\percent} accuracy mark (\cref{fig:control}a).
For the \gls{sgoogle}, overfitting was less pronounced, but also the overall test accuracy dropped to \SI{20}{\percent} (\cref{fig:control}b).
Thus linear classifiers provided a low degree of generalization.

To assess whether this situation was different for nonlinear classifiers, we trained \glspl{svm} with polynomial kernels up to a degree of~3.
For these kernels, overfitting was less pronounced.
Slightly better performance of about \SI{60}{\percent} on the \gls{sxxx} and \SI{30}{\percent} on the \gls{sgoogle} was achieved when using a \gls{svm} with a \gls{rbf} kernel.
The performance on the \gls{sxxx} test set, which includes speakers that are not
part of the training set, was noticeably lower compared to the accuracy on the validation data.
Especially, for polynomial and \gls{rbf} kernels the generalization across speakers was worse than for the linear kernel (\cref{fig:control}a).
In contrast, we found the performance on the \gls{sgoogle} test set to be on par
with the accuracy on the validation set (\cref{fig:control}b), which is most
likely an effect of the uniform speaker distribution.
These results illustrate that both linear and nonlinear classifiers trained on
spike count patterns without temporal information were unable to surpass the
\SI{60}{\percent} accuracy mark for the \gls{sxxx} and the \SI{30}{\percent} mark for the \gls{sgoogle} dataset.
Therefore, spike counts are not sufficient to achieve high classification accuracy on the studied datasets.

\begin{figure*}[t]
	\begin{minipage}{0.55\textwidth}
		\centering
		\begin{tikzpicture}
			\node (figure) {\input{graphics/loss}};
			\node[draw, fill=white] (ff) at (1.8, 1.95) {\scalebox{0.4}{\input{graphics/networks}}};
		\end{tikzpicture}
	\end{minipage}
	\hfill
	\begin{minipage}{0.40\textwidth}
		\centering
		\begin{tikzpicture}
			\node (figure){\input{figures/loss.pgf}};
			\draw[dashed] (2.2,-2.6) -- (2.2,2.6);
		\end{tikzpicture}
	\end{minipage}
	\caption{%
		\textbf{Setup and multiple choices of loss functions for \glspl{snn}.} 
		\textbf{(a)}~Schematic of a single layer recurrent network with two readout units.
		We applied two different loss functions for \glspl{lstm} and \glspl{snn}:
		First, a max-over-time loss was considered, where the time step with maximal activity of each readout was used to calculate the cross entropy (marked by colored arrows).
		Second, a last-time-step loss was utilized where only the last time step of the activation was considered in the calculation of the cross entropy (marked by gray arrow).
		The inset illustrates the corresponding feed-forward topology.
		\textbf{(b)}~Bar graph of classification accuracy for different \acrshortpl{snn} and a \acrshort{lstm} on the \acrshort{sxxx}.
		Only the \acrshort{lstm} generalized well when trained with a last-time-step loss.
		\textbf{(c)}~Same as in (b), but showing performance for a max-over-time loss.
		Overall, \acrshortpl{snn} and \acrshortpl{lstm} performed better when trained with the max-over-time loss.
	}
	\label{fig:loss_schematic}
\end{figure*}
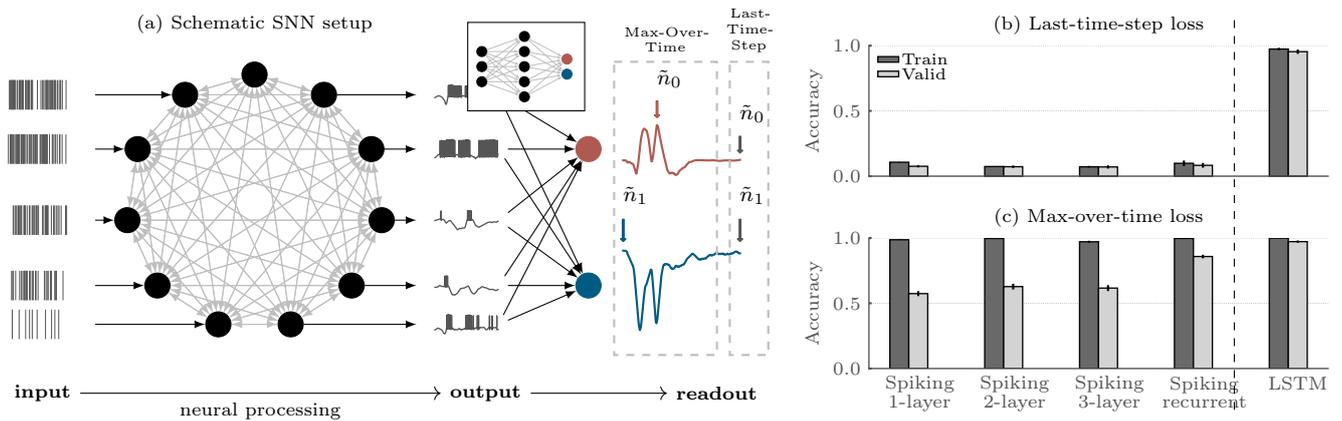

Next, we wanted to assess whether decoding accuracy could be improved when training classifiers that have explicit access to temporal information of the spike times.
Therefore, we trained \glspl{lstm} on temporal histograms of spiking activity
(\cref{subsubsec:LSTM}).
In spite of the small size of the \gls{sxxx} dataset, \glspl{lstm} showed
reduced overfitting and were able to solve the classification problem with an
accuracy of \SI{85.7(14)}{\percent} (\cref{fig:control}a) which was
substantially higher than the best performing \gls{svm}.
Similarly, for the \gls{sgoogle} dataset the \gls{lstm} test accuracy \SI{75.0(2)}{\percent} was more than twice as high as the best-performing classifier on the spike count data.
However, the degree of overfitting was slightly higher than on \gls{sxxx}.

Since both, kernel machines and \glspl{lstm}, were affected by overfitting, we
tested whether performance could be increased with \glspl{cnn} due to their
inductive bias on translation invariance in both frequency and time and their reduced number of parameters. 
To that end, we binned spikes in spatio-temporal histograms and trained a
\gls{cnn} classifier (\cref{subsubsec:CNN}).
\Glspl{cnn} showed the least amount of overfitting among all tested classifiers; the accuracy dropped by only \SI{1.4}{\percent} on \gls{sxxx} and by \SI{1.5}{\percent} on \gls{sgoogle} (\cref{fig:control}). 
Especially, the performance on the \gls{sxxx} test data was on par with the one
on the validation set, demonstrating a high degree of generalisation.

These findings highlight that the temporal information contained in both datasets can be exploited by suitable neural network architectures. 
Moreover, these results provide a lower bound on the performance ceiling for both datasets. 
It seems likely that a more careful architecture search and hyperparameter tuning will only improve upon these results.
Thus, both the \gls{sxxx} and the \gls{sgoogle} will be useful for quantitative comparison between \glspl{snn} up to at least these empirical accuracy values.

\subsection{Training spiking neural networks}

Having established that both spiking datasets contain useful temporal
information that can be read out by a suitable classifier, 
we sought to train \glspl{snn} of \gls{lif} neurons using \gls{bptt} to
establish a first set of baselines and to assess their generalization
properties. 
One problem with training \glspl{snn} with gradient descent arises because the derivative of the neural activation function appears in the evaluation of the gradient.
Since spiking is an intrinsically discontinuous process, the resulting gradients are ill-defined.
To nevertheless train networks of \gls{lif} neurons using supervised loss
functions, we used a surrogate gradient approach~\cite{neftci_surrogate_2019}.
Surrogate gradients can be seen as a continuous relaxation of the real gradients
of a \gls{snn} which can be implemented as an in-place replacement while performing \gls{bptt}.
Importantly, we did not change the neuron model and the associated forward-pass of the model, but used a fast sigmoid as a surrogate activation function when computing gradients (Methods \cref{subsubsec:supervised}).

Although not a requirement~\cite{neftci_surrogate_2019,li_learning_2020}, we only considered \glspl{snn} with fixed, finite time constants on the order of \SI{}{\milli\second} inspired by biology.
Because of this constraint, we investigate two different loss functions for both \glspl{lstm} and \glspl{snn} (\cref{fig:loss_schematic}a).
The results for \glspl{lstm} shown in \cref{fig:control} were obtained by training with a \textit{last-time-step loss}, where the activation of the last time step of each example and readout unit was used to calculate the cross entropy loss at the output.
In addition, we also considered a \textit{max-over-time loss}, in which the time
step with maximum activation of each readout unit was taken into account (\cref{fig:loss_schematic}a).
This loss function is motivated by the Tempotron~\cite{gutig_tempotron:_2006} in which the network signals its decision about the class membership of the applied input pattern by whether a neuron spiked or not. 

We evaluated the performance of \glspl{lstm} and \glspl{snn} for both aforementioned loss functions on the \gls{sxxx}.
Training \glspl{lstm} with a cross entropy loss based on the activity of the last time step of every sample was associated with high performance in contrast to \glspl{snn} (\cref{fig:loss_schematic}b).
The slightly reduced performance of feed-forward \glspl{snn} trained with last-time-step loss compared to \glspl{rsnn} suggests that time constants were too low to provide all necessary information at the last time step.
This was presumably due to active memory implemented through reverberating activity through the recurrent connections.
Overall, \glspl{snn} performed better in combination with the max-over-time loss function (\cref{fig:loss_schematic}c).
Also \glspl{lstm} showed increased performance in combination with a max-over-time loss; the validation accuracy increased from \SI{95.4(17)}{\percent} for the last-time-step loss to \SI{97.2(09)}{\percent} for the max-over-time loss.
Motivated by these results, we used a max-over-time loss for \glspl{snn} as well as \glspl{lstm} throughout the remainder of this manuscript.

\begin{figure}[t]
	\centering
	\input{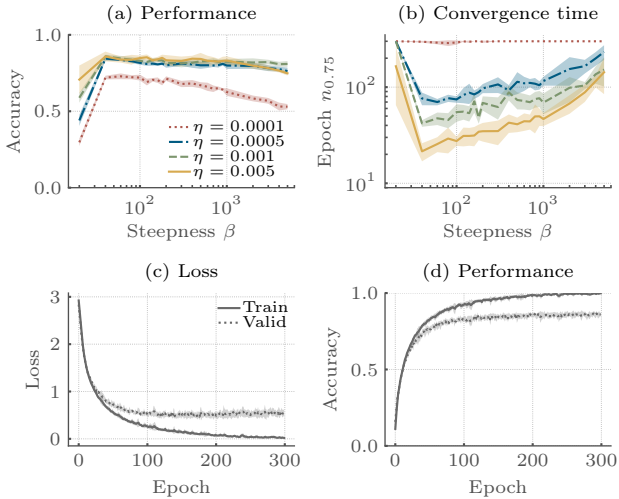}
	\caption{%
		\textbf{Accuracy, but not convergence time is only mildly affected by the steepness $\beta$ of the surrogate derivative.}
		\textbf{(a)}~Accuracy as a function of $\beta$ on a validation set of the \acrshort{sxxx} for different learning rates $\eta$.
		Performance is highest for a wide range of $\beta$ values ($\beta \ge 40$) and depends only slightly on $\eta$.
		\textbf{(b)}~Number of epochs needed to reach an accuracy $> 0.75$, $n_{0.75}$.
		In contrast to the performance, $n_{0.75}$ strongly depends on both $\beta$ and $\eta$.
		\textbf{(c)}~Loss curves on the \acrshort{sxxx} for $\beta = 40$ and $\eta = 10^{-3}$.
		\textbf{(d)}~Same as in (c), but showing the accuracy on the \acrshortpl{sxxx}.
	}
	\label{fig:parameter}
\end{figure}

Surrogate gradient learning introduces a new hyperparameter $\beta$ associated with the steepness of the surrogate derivative (Methods \cref{eq:surrogate_gradient}).
Because changes in $\beta$ may require a different optimal learning rate $\eta$,
we performed a grid search over $\beta$ and $\eta$  based on a single-layer \gls{rsnn} architecture trained on \gls{sxxx}.
We found that sensible combinations for both parameters lead to stable performance plateaus over a large range of values (\cref{fig:parameter}a).
Only for small $\beta$ the accuracy dropped dramatically, whereas it decreased only slowly for high values.
Interestingly, the learning rate had hardly any effect on peak performance for the tested parameter values.
As expected, convergence speed heavily depended on both $\eta$ and $\beta$.
These results motivated us to use $\beta = \SI{40}{}$ and $\eta = \SI{1e-3}{}$ for all \glspl{snn} architectures presented in this article unless mentioned otherwise.
For this choice, the performance of the \gls{rsnn} on the validation set reached its peak after about 150 epochs (\cref{fig:parameter}d).
Additional training only increased performance on the training dataset (\cref{fig:parameter}c), but did not impact generalization (\cref{fig:parameter}d).

\begin{figure}[t]
	\centering
	\begin{tikzpicture}
		\node (figure){\input{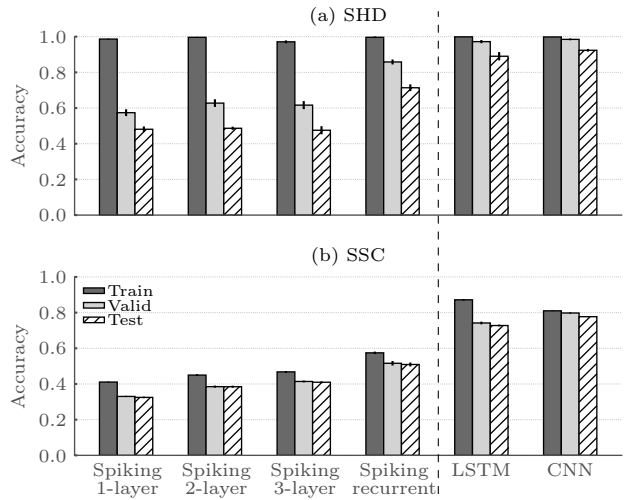}};
		\draw[dashed] (1.6,-3.2) -- (1.6,3.2);
	\end{tikzpicture}
	\caption{%
		\textbf{Recurrent \acrshortpl{snn} outperform feed-forward architectures on both datasets.}
		\textbf{(a)}~Bar graph of classification accuracy for different \acrshort{snn} architectures on the \acrshort{sxxx}.
		The accuracy reached by \acrshort{rsnn} is comparable to the performance of \acrshortpl{lstm} with a max-over-time loss.
		Increasing the number of layers in feed-forward architectures hardly affected performance.
		\textbf{(b)}~Same as in (a), but showing performance on the \acrshort{sgoogle}.
		The performance of \acrshortpl{snn} was lower than the one reached by \acrshortpl{lstm}.
		In contrast to (a), an increasing number of layers lead to a monotonic increase of accuracy.
	}
	\label{fig:spiking}
\end{figure}

With the parameter choices discussed above, we trained various \gls{snn} architectures on the \gls{sxxx} and the \gls{sgoogle}.
To that end, we considered multi-layer feed-forward \glspl{snn} with $l$ layers and a single-layer \gls{rsnn}.
Interestingly, increasing the number of hidden layers $l$ did not notably improve performance on the \gls{sxxx} (\cref{fig:spiking}a).
In addition, all choices of $l$ caused high levels of overfitting.
Moreover, feed-forward \glspl{snn} reached slightly lower accuracy levels than the \glspl{svm} on the \gls{sxxx} (\cref{fig:control}a).
For the larger \gls{sgoogle} dataset, the degree of overfitting was much smaller
(\cref{fig:spiking}b) and performance was markedly better than the one reached by \glspl{svm} (\cref{fig:control}b).
Here, increasing the number of layers of feed-forward \glspl{snn} led to a
monotonic increase of performance on the test set from \SI{32.5(5)}{\percent}
for a single layer to \SI{41.0(5)}{\percent} in the three-layer case ($l=3$).
However, when testing \glspl{rsnn}, we found consistently higher performance and improved generalization across speakers.
In comparison to the accuracy of \glspl{lstm}, \glspl{rsnn} showed higher overfitting, and generalized less well across speakers.
The \gls{rsnn} achieved the highest accuracy of \SI{71.4(19)}{\percent} on the \gls{sxxx} and \SI{50.9(11)}{\percent} on the \gls{sgoogle} which was still less than the \gls{lstm} with \SI{85.7(14)}{\percent} on the \gls{sxxx} and \SI{75.0(2)}{\percent} on the \gls{sgoogle}.

\subsection{Generalization across speakers and datasets}

\begin{figure}[t]
	\centering
	\input{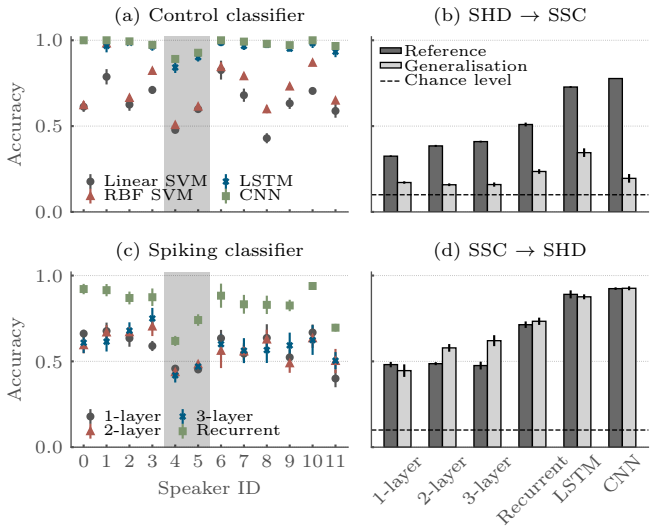}
	\caption{%
		\textbf{Networks generalize across speakers and datasets.}
		By reserving two speakers for the test set, the \acrshort{sxxx} dataset allows to assess speaker generalization performance.
		\textbf{(a)}~Per-speaker classification accuracy on the test set of the \acrshort{sxxx} for different classifiers.
		A clear decrease in performance was observable for samples spoken by the held-out speakers four and five (highlighted).
		\textbf{(b)}~Bar graph of the performance of \acrshortpl{snn}, \acrshort{lstm} and \acrshort{cnn} trained on the \acrshort{sxxx} and tested on the English digits of the \acrshort{sgoogle}.
		The reference is given by the performance on the \acrshort{sxxx} test set, and the generalisation by the performance on the English digits of the \acrshort{sgoogle} test set.
		Accuracy on the \acrshort{sgoogle} digits was substantially lower than on the digits in the \acrshort{sgoogle} test set.
		\textbf{(c)}~Same as in (a), but showing the per-speaker accuracy of \acrshortpl{snn}.
		As for (a), a decrease in performance for the held-out speakers was observed.
		\textbf{(d)}~Same as in (b), but showing the performance of networks when trained on \acrshort{sgoogle} and tested on the English digits of the \acrshort{sxxx}.
		As opposed to (b), networks trained on the \acrshort{sgoogle} digits generalize well across datasets.
	}
	\label{fig:generalization}
\end{figure}

For robust spoken word classification, the generalization across speakers is a key feature.
This generalization can be assessed by evaluating the accuracy per speaker on \gls{sxxx}, as the digits spoken by speakers four and five are only present in the test set.
We compared the performance on the digits of the held-out speakers to all other speakers and found a clear performance drop across all classification methods for the speakers four and five (\cref{fig:generalization}a and c).
For \glspl{svm}, the linear kernel led to the smallest accuracy drop of about \SI{18}{\percent}, whereas we found a decrease of \SI{26}{\percent} for the \gls{rbf} kernel.
\Glspl{cnn} generalized best with a drop of only \SI{8}{\percent}, followed by the \glspl{lstm} with \SI{10}{\percent}.
Among \glspl{snn}, feed-forward architectures were most strongly affected with a drop of about \SIrange{24}{27}{\percent}.
\Glspl{rsnn}, however, only underwent a decline of \SI{21}{\percent} in performance (\cref{fig:generalization}b).
This illustrates that the composition of the test set of \gls{sxxx} can provide meaningful information with regard to  generalization across speakers. 

Because English digits are part of both datasets, we were able to test the generalization across datasets by training \glspl{snn}, \gls{lstm} and \glspl{cnn} on the full \gls{sxxx} dataset while testing on a restricted \gls{sgoogle} dataset and vice versa (\cref{fig:generalization}b and d).
For testing, the datasets were restricted to the common English digits \textit{zero} to \textit{nine}.
Perhaps not surprisingly, networks generalized better, when trained on the larger \gls{sgoogle} dataset as a reference and tested on \gls{sxxx}.
Nevertheless, all architectures trained on the \gls{sxxx} and tested on the \gls{sgoogle} reached performance levels above chance.
Again, recurrent architectures reached highest performance among all tested \glspl{snn}.

\subsection{Improving generalization performance through data augmentation and larger networks}

\begin{figure}[t]
	\centering
	\input{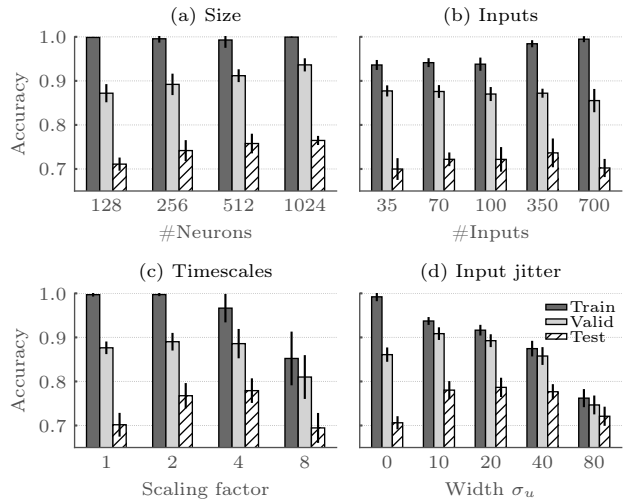}
	\caption{%
		\textbf{Generalization performance improves with network size and data augmentation.}
		\textbf{(a)}~Bar graph of classification accuracy for different network sizes.
		Larger networks generalize best.
		\textbf{(b)}~Same as before, but showing the accuracy for different numbers of input channels.
		Networks generalized best when the spikes of the \SI{700}{} input units	were compressed to \SI{70}{} units.
		\textbf{(c)}~Network performance for temporally scaled synaptic and membrane time constants.
		Highest performance on the validation set was reached for an expansive scaling factor of \SI{4}{}.
		\textbf{(d)}~Classification accuracy for spatially noisy (channel jitter) input spikes.
		An optimum occurs for a standard deviation of $\sigma_u = \SI{20}{}$ channels. 
	}
	\label{fig:best}
\end{figure}

We first tested the effect of network size on the generalization performance of the \gls{rsnn} architecture using the smaller \gls{sxxx} dataset.
As expected, we found that increasing the network size up to \SI{1024}{} neurons indeed resulted in improved validation and test accuracy up to \SI{76.5(10)}{\percent} with \SI{1024}{} neurons 
 (\cref{fig:best}a).

Next, we compressed the input in terms of channels, since a similar modification led to good results for the \glspl{cnn} (cf.\ \cref{fig:spiking}).
To that end, we coarse-grained the inputs by condensing the spike trains of neighboring input units.
Merging \SI{10}{} neighboring channels channels did improve validation accuracy with an associated test accuracy of \SI{72.2(16)}{\percent} (\cref{fig:best}b).
This choice resulted in an overall channel count of \SI{70}{}.
Despite the only slight increase in test accuracy, the corresponding networks with fewer channels benefited from a smaller computational footprint.

We next studied whether compressing the time domain would bear additional benefits.
In more detail, we rescaled all neuronal time constants in the network.
For a scaling factor of \SI{4}{} corresponding to $\tau_\mathrm{mem} = \SI{80}{\milli\second}$ and $\tau_\mathrm{syn} = \SI{40}{\milli\second}$, the manipulation led to a marked reduction of overfitting that also resulted in a higher test accuracy of \SI{79.9(28)}{\percent} (\cref{fig:best}c).

As an alternative means to reduce overfitting, we sought to explored noise injection at the input layer.
Specifically, we implemented event-based spike jitter across channels by adding a random number $\mathcal{N}(0, \sigma_u)$ to the unit index $i$ of every input spike, which was then rounded to the nearest integer.
We found that such input noise was effective in decreasing overfitting.
For instance, a value $\sigma_u = \SI{20}{}$ led to an increased test accuracy of \SI{78.7(22)}{\percent} (\cref{fig:best}d).

When combined, the aforementioned strategies resulted in a further improved \textit{best effort} test accuracy of \SI{83.2(13)}{\percent} on the \gls{sxxx} test set (\cref{tab:performance}).
This is not only noticeably higher than the previous result of \SI{71.4(19)}{\percent} obtained by the \gls{rsnn}, but also exceeds the effect of every contribution on its own.

\begin{table}[tb]
	\centering
	\TableFontSize
	\begin{threeparttable}
		\caption{%
			Performance comparison
		}
		\label{tab:performance}
		\input{tab/performance}
	\end{threeparttable}
\end{table}

Finally, to relate our \gls{lif} neurons to other work on \glspl{rsnn}, we also
evaluated the performance of the \glspl{snu}~\cite{wozniak_deep_2020} on \gls{sxxx}.
Although closely related to our \gls{lif} neuron model, \glspl{snu} feature
delta synapses and rely on a different surrogate derivative.
When trained under the same conditions as our \textit{best effort} network, the
\gls{snu} network reached a test accuracy of \SI{79.0(16)}{\percent} (\cref{tab:performance})
which is slightly lower than networks using our \gls{lif} neuron model.

%% file: graphics/loss.tex
\begin{tikzpicture}[x=1.3cm, y=1.3cm]
	\pgfmathsetmacro{\n}{10}  
	\pgfmathsetmacro{\r}{1.3} 
	\definecolor{patterncolor}{RGB}{0,0,0} 
	\definecolor{nicered}{HTML}{AF5A50}
	\definecolor{niceblue}{HTML}{005B82}
	
	\coordinate (center) at ({5.0*\r},0);

	\foreach \i in {6,...,10}{
		\node (input\i) at ($({-1.65*\r},{\r*cos(\i*360/(\n+1))})+(center)$) {\input{figures/input_\i.pgf}};
	}

	\foreach \i in {0,1,...,\n}{
		\node[circle,draw,patterncolor,fill] (hidden\i) at ($({\r*sin(\i*360/(\n+1))},{\r*cos(\i*360/(\n+1))})+(center)$) {};
	}

	\foreach \i in {1,2,...,5}{
		\node (output\i) at ($({1.65*\r},{\r*cos(\i*360/(\n+1))})+(center)$) {\input{figures/hidden_\i.pgf}};
	}
	
	\foreach \i in {1,2,...,5}{
		\node (tempoutput\i) at ($({1.9*\r},{\r*cos(\i*360/(\n+1))})+(center)$) {};
	}

	\node[circle,draw,nicered,fill] (readoutunit2) at ($({2.6*\r},{\r*cos(2*360/(\n+1))})+(center)$) {};
	\node[circle,draw,niceblue,fill] (readoutunit4) at ($({2.6*\r},{\r*cos(4*360/(\n+1))})+(center)$) {};

	\foreach \i in {2,4}{
		\node (readout\i) at ($({3.4*\r},{\r*cos(\i*360/(\n+1))})+(center)+(0,0.20*\r)$) {\input{figures/readout_\i.pgf}};
	}

	\foreach \i in {6,...,10}{
		\draw[-latex,patterncolor] (input\i) to (hidden\i);
	}

  	\foreach \source in {0,...,\n}{
		\foreach \dest in {0,...,\n}{
			\ifthenelse{\equal{\source}{\dest}}{
			}{
				\draw[-latex,lightgray] (hidden\source) to (hidden\dest);
			}
		}
	}

	\foreach \i in {1,...,5}{
		\draw[-latex,patterncolor] (hidden\i) to (output\i);
	}

	\foreach \source in {1,...,5}{
		\foreach \dest in {2,4}{
			\draw[-latex,patterncolor] (tempoutput\source) to (readoutunit\dest);
		}
	}

	\node[font=\scriptsize\bfseries] (label1) at ($(-1.65*\r,-1.5*\r)+(center)$) {input};
	\node[font=\scriptsize\bfseries] (label2) at ($(1.8*\r,-1.5*\r)+(center)$) {output};
	\node[font=\scriptsize\bfseries] (label3) at ($(3.6*\r,-1.5*\r)+(center)$) {readout};
	\node[font=\scriptsize] (title) at ($(0*\r, 1.4*\r)+(center)$) {(a) Schematic SNN setup};
	\draw[->] (label1) to node[below] {\scriptsize{neural processing}} (label2);
	\draw[->] (label2) to node[below] {} (label3);

	\draw[thick,dashed,lightgray] ($({3.7*\r},{1.1*\r})+(center)$)
		to node[above,text width=0.8cm,font=\tiny,align=center,text=black] {Last-Time-Step} ($({4.0*\r},{1.1*\r})+(center)$)
		to ($({4.0*\r},{-1.2*\r})+(center)$)
		to ($({3.7*\r},{-1.2*\r})+(center)$)
		to ($({3.7*\r},{1.1*\r})+(center)$);
	\draw[thick,dashed,lightgray] ($({2.8*\r},{1.1*\r})+(center)$)
		to node[above,text width=1.1cm,font=\tiny,align=center,text=black] {Max-Over-Time} ($({3.6*\r},{1.1*\r})+(center)$)
		to ($({3.6*\r},{-1.2*\r})+(center)$)
		to ($({2.8*\r},{-1.2*\r})+(center)$)
		to ($({2.8*\r},{1.1*\r})+(center)$);

\end{tikzpicture}

%% file: graphics/networks.tex
\begin{tikzpicture}[x=0.7cm, y=0.5cm]
	\definecolor{patterncolor}{RGB}{0,0,0}
	\definecolor{nicered}{HTML}{AF5A50}
	\definecolor{niceblue}{HTML}{005B82}
	
	\foreach \i in {1,...,3}{
		\node[circle, draw,patterncolor,fill] (input\i) at (0, \i) {};
	}
	
	\foreach \i in {1,...,5}{
		\node[circle,draw,patterncolor,fill] (hidden\i) at (2,\i-1) {};
	}
	
	\foreach \i in {1,...,3}{
		\foreach \j in {1,...,5}{
			\draw[-latex,lightgray] (input\i) to (hidden\j);
		}
	}
	
	\node[circle,draw,niceblue,fill] (readout1) at (4,1.5) {};
	\node[circle,draw,nicered,fill] (readout2) at (4,2.5) {};
	
	\foreach \i in {1,...,5}{
		\foreach \j in {1,...,2}{
			\draw[-latex,lightgray] (hidden\i) to (readout\j);
		}
	}
\end{tikzpicture}

%% file: figures/loss.pgf
\begingroup%
\makeatletter%
\begin{pgfpicture}%
\pgfpathrectangle{\pgfpointorigin}{\pgfqpoint{2.744637in}{2.100000in}}%
\pgfusepath{use as bounding box, clip}%
\begin{pgfscope}%
\pgfsetbuttcap%
\pgfsetmiterjoin%
\pgfsetlinewidth{0.000000pt}%
\definecolor{currentstroke}{rgb}{0.000000,0.000000,0.000000}%
\pgfsetstrokecolor{currentstroke}%
\pgfsetstrokeopacity{0.000000}%
\pgfsetdash{}{0pt}%
\pgfpathmoveto{\pgfqpoint{0.000000in}{0.000000in}}%
\pgfpathlineto{\pgfqpoint{2.744637in}{0.000000in}}%
\pgfpathlineto{\pgfqpoint{2.744637in}{2.100000in}}%
\pgfpathlineto{\pgfqpoint{0.000000in}{2.100000in}}%
\pgfpathclose%
\pgfusepath{}%
\end{pgfscope}%
\begin{pgfscope}%
\pgfsetbuttcap%
\pgfsetmiterjoin%
\pgfsetlinewidth{0.000000pt}%
\definecolor{currentstroke}{rgb}{0.000000,0.000000,0.000000}%
\pgfsetstrokecolor{currentstroke}%
\pgfsetstrokeopacity{0.000000}%
\pgfsetdash{}{0pt}%
\pgfpathmoveto{\pgfqpoint{0.345807in}{1.245917in}}%
\pgfpathlineto{\pgfqpoint{2.724637in}{1.245917in}}%
\pgfpathlineto{\pgfqpoint{2.724637in}{1.927250in}}%
\pgfpathlineto{\pgfqpoint{0.345807in}{1.927250in}}%
\pgfpathclose%
\pgfusepath{}%
\end{pgfscope}%
\begin{pgfscope}%
\pgfsetbuttcap%
\pgfsetroundjoin%
\definecolor{currentfill}{rgb}{0.317647,0.317647,0.317647}%
\pgfsetfillcolor{currentfill}%
\pgfsetlinewidth{0.803000pt}%
\definecolor{currentstroke}{rgb}{0.317647,0.317647,0.317647}%
\pgfsetstrokecolor{currentstroke}%
\pgfsetdash{}{0pt}%
\pgfsys@defobject{currentmarker}{\pgfqpoint{0.000000in}{0.000000in}}{\pgfqpoint{0.000000in}{0.013889in}}{%
\pgfpathmoveto{\pgfqpoint{0.000000in}{0.000000in}}%
\pgfpathlineto{\pgfqpoint{0.000000in}{0.013889in}}%
\pgfusepath{stroke,fill}%
}%
\begin{pgfscope}%
\pgfsys@transformshift{0.503085in}{1.245917in}%
\pgfsys@useobject{currentmarker}{}%
\end{pgfscope}%
\end{pgfscope}%
\begin{pgfscope}%
\pgfsetbuttcap%
\pgfsetroundjoin%
\definecolor{currentfill}{rgb}{0.317647,0.317647,0.317647}%
\pgfsetfillcolor{currentfill}%
\pgfsetlinewidth{0.803000pt}%
\definecolor{currentstroke}{rgb}{0.317647,0.317647,0.317647}%
\pgfsetstrokecolor{currentstroke}%
\pgfsetdash{}{0pt}%
\pgfsys@defobject{currentmarker}{\pgfqpoint{0.000000in}{0.000000in}}{\pgfqpoint{0.000000in}{0.013889in}}{%
\pgfpathmoveto{\pgfqpoint{0.000000in}{0.000000in}}%
\pgfpathlineto{\pgfqpoint{0.000000in}{0.013889in}}%
\pgfusepath{stroke,fill}%
}%
\begin{pgfscope}%
\pgfsys@transformshift{0.994579in}{1.245917in}%
\pgfsys@useobject{currentmarker}{}%
\end{pgfscope}%
\end{pgfscope}%
\begin{pgfscope}%
\pgfsetbuttcap%
\pgfsetroundjoin%
\definecolor{currentfill}{rgb}{0.317647,0.317647,0.317647}%
\pgfsetfillcolor{currentfill}%
\pgfsetlinewidth{0.803000pt}%
\definecolor{currentstroke}{rgb}{0.317647,0.317647,0.317647}%
\pgfsetstrokecolor{currentstroke}%
\pgfsetdash{}{0pt}%
\pgfsys@defobject{currentmarker}{\pgfqpoint{0.000000in}{0.000000in}}{\pgfqpoint{0.000000in}{0.013889in}}{%
\pgfpathmoveto{\pgfqpoint{0.000000in}{0.000000in}}%
\pgfpathlineto{\pgfqpoint{0.000000in}{0.013889in}}%
\pgfusepath{stroke,fill}%
}%
\begin{pgfscope}%
\pgfsys@transformshift{1.486073in}{1.245917in}%
\pgfsys@useobject{currentmarker}{}%
\end{pgfscope}%
\end{pgfscope}%
\begin{pgfscope}%
\pgfsetbuttcap%
\pgfsetroundjoin%
\definecolor{currentfill}{rgb}{0.317647,0.317647,0.317647}%
\pgfsetfillcolor{currentfill}%
\pgfsetlinewidth{0.803000pt}%
\definecolor{currentstroke}{rgb}{0.317647,0.317647,0.317647}%
\pgfsetstrokecolor{currentstroke}%
\pgfsetdash{}{0pt}%
\pgfsys@defobject{currentmarker}{\pgfqpoint{0.000000in}{0.000000in}}{\pgfqpoint{0.000000in}{0.013889in}}{%
\pgfpathmoveto{\pgfqpoint{0.000000in}{0.000000in}}%
\pgfpathlineto{\pgfqpoint{0.000000in}{0.013889in}}%
\pgfusepath{stroke,fill}%
}%
\begin{pgfscope}%
\pgfsys@transformshift{1.977567in}{1.245917in}%
\pgfsys@useobject{currentmarker}{}%
\end{pgfscope}%
\end{pgfscope}%
\begin{pgfscope}%
\pgfsetbuttcap%
\pgfsetroundjoin%
\definecolor{currentfill}{rgb}{0.317647,0.317647,0.317647}%
\pgfsetfillcolor{currentfill}%
\pgfsetlinewidth{0.803000pt}%
\definecolor{currentstroke}{rgb}{0.317647,0.317647,0.317647}%
\pgfsetstrokecolor{currentstroke}%
\pgfsetdash{}{0pt}%
\pgfsys@defobject{currentmarker}{\pgfqpoint{0.000000in}{0.000000in}}{\pgfqpoint{0.000000in}{0.013889in}}{%
\pgfpathmoveto{\pgfqpoint{0.000000in}{0.000000in}}%
\pgfpathlineto{\pgfqpoint{0.000000in}{0.013889in}}%
\pgfusepath{stroke,fill}%
}%
\begin{pgfscope}%
\pgfsys@transformshift{2.469060in}{1.245917in}%
\pgfsys@useobject{currentmarker}{}%
\end{pgfscope}%
\end{pgfscope}%
\begin{pgfscope}%
\pgfpathrectangle{\pgfqpoint{0.345807in}{1.245917in}}{\pgfqpoint{2.378830in}{0.681333in}}%
\pgfusepath{clip}%
\pgfsetbuttcap%
\pgfsetroundjoin%
\pgfsetlinewidth{0.301125pt}%
\definecolor{currentstroke}{rgb}{0.725490,0.725490,0.725490}%
\pgfsetstrokecolor{currentstroke}%
\pgfsetdash{{0.300000pt}{0.495000pt}}{0.000000pt}%
\pgfpathmoveto{\pgfqpoint{0.345807in}{1.245917in}}%
\pgfpathlineto{\pgfqpoint{2.724637in}{1.245917in}}%
\pgfusepath{stroke}%
\end{pgfscope}%
\begin{pgfscope}%
\pgfsetbuttcap%
\pgfsetroundjoin%
\definecolor{currentfill}{rgb}{0.317647,0.317647,0.317647}%
\pgfsetfillcolor{currentfill}%
\pgfsetlinewidth{0.803000pt}%
\definecolor{currentstroke}{rgb}{0.317647,0.317647,0.317647}%
\pgfsetstrokecolor{currentstroke}%
\pgfsetdash{}{0pt}%
\pgfsys@defobject{currentmarker}{\pgfqpoint{0.000000in}{0.000000in}}{\pgfqpoint{0.013889in}{0.000000in}}{%
\pgfpathmoveto{\pgfqpoint{0.000000in}{0.000000in}}%
\pgfpathlineto{\pgfqpoint{0.013889in}{0.000000in}}%
\pgfusepath{stroke,fill}%
}%
\begin{pgfscope}%
\pgfsys@transformshift{0.345807in}{1.245917in}%
\pgfsys@useobject{currentmarker}{}%
\end{pgfscope}%
\end{pgfscope}%
\begin{pgfscope}%
\definecolor{textcolor}{rgb}{0.317647,0.317647,0.317647}%
\pgfsetstrokecolor{textcolor}%
\pgfsetfillcolor{textcolor}%
\pgftext[x=0.160429in,y=1.213800in,left,base]{\color{textcolor}\rmfamily\fontsize{6.664000}{7.996800}\selectfont \(\displaystyle 0.0\)}%
\end{pgfscope}%
\begin{pgfscope}%
\pgfpathrectangle{\pgfqpoint{0.345807in}{1.245917in}}{\pgfqpoint{2.378830in}{0.681333in}}%
\pgfusepath{clip}%
\pgfsetbuttcap%
\pgfsetroundjoin%
\pgfsetlinewidth{0.301125pt}%
\definecolor{currentstroke}{rgb}{0.725490,0.725490,0.725490}%
\pgfsetstrokecolor{currentstroke}%
\pgfsetdash{{0.300000pt}{0.495000pt}}{0.000000pt}%
\pgfpathmoveto{\pgfqpoint{0.345807in}{1.586583in}}%
\pgfpathlineto{\pgfqpoint{2.724637in}{1.586583in}}%
\pgfusepath{stroke}%
\end{pgfscope}%
\begin{pgfscope}%
\pgfsetbuttcap%
\pgfsetroundjoin%
\definecolor{currentfill}{rgb}{0.317647,0.317647,0.317647}%
\pgfsetfillcolor{currentfill}%
\pgfsetlinewidth{0.803000pt}%
\definecolor{currentstroke}{rgb}{0.317647,0.317647,0.317647}%
\pgfsetstrokecolor{currentstroke}%
\pgfsetdash{}{0pt}%
\pgfsys@defobject{currentmarker}{\pgfqpoint{0.000000in}{0.000000in}}{\pgfqpoint{0.013889in}{0.000000in}}{%
\pgfpathmoveto{\pgfqpoint{0.000000in}{0.000000in}}%
\pgfpathlineto{\pgfqpoint{0.013889in}{0.000000in}}%
\pgfusepath{stroke,fill}%
}%
\begin{pgfscope}%
\pgfsys@transformshift{0.345807in}{1.586583in}%
\pgfsys@useobject{currentmarker}{}%
\end{pgfscope}%
\end{pgfscope}%
\begin{pgfscope}%
\definecolor{textcolor}{rgb}{0.317647,0.317647,0.317647}%
\pgfsetstrokecolor{textcolor}%
\pgfsetfillcolor{textcolor}%
\pgftext[x=0.160429in,y=1.554467in,left,base]{\color{textcolor}\rmfamily\fontsize{6.664000}{7.996800}\selectfont \(\displaystyle 0.5\)}%
\end{pgfscope}%
\begin{pgfscope}%
\pgfpathrectangle{\pgfqpoint{0.345807in}{1.245917in}}{\pgfqpoint{2.378830in}{0.681333in}}%
\pgfusepath{clip}%
\pgfsetbuttcap%
\pgfsetroundjoin%
\pgfsetlinewidth{0.301125pt}%
\definecolor{currentstroke}{rgb}{0.725490,0.725490,0.725490}%
\pgfsetstrokecolor{currentstroke}%
\pgfsetdash{{0.300000pt}{0.495000pt}}{0.000000pt}%
\pgfpathmoveto{\pgfqpoint{0.345807in}{1.927250in}}%
\pgfpathlineto{\pgfqpoint{2.724637in}{1.927250in}}%
\pgfusepath{stroke}%
\end{pgfscope}%
\begin{pgfscope}%
\pgfsetbuttcap%
\pgfsetroundjoin%
\definecolor{currentfill}{rgb}{0.317647,0.317647,0.317647}%
\pgfsetfillcolor{currentfill}%
\pgfsetlinewidth{0.803000pt}%
\definecolor{currentstroke}{rgb}{0.317647,0.317647,0.317647}%
\pgfsetstrokecolor{currentstroke}%
\pgfsetdash{}{0pt}%
\pgfsys@defobject{currentmarker}{\pgfqpoint{0.000000in}{0.000000in}}{\pgfqpoint{0.013889in}{0.000000in}}{%
\pgfpathmoveto{\pgfqpoint{0.000000in}{0.000000in}}%
\pgfpathlineto{\pgfqpoint{0.013889in}{0.000000in}}%
\pgfusepath{stroke,fill}%
}%
\begin{pgfscope}%
\pgfsys@transformshift{0.345807in}{1.927250in}%
\pgfsys@useobject{currentmarker}{}%
\end{pgfscope}%
\end{pgfscope}%
\begin{pgfscope}%
\definecolor{textcolor}{rgb}{0.317647,0.317647,0.317647}%
\pgfsetstrokecolor{textcolor}%
\pgfsetfillcolor{textcolor}%
\pgftext[x=0.160429in,y=1.895133in,left,base]{\color{textcolor}\rmfamily\fontsize{6.664000}{7.996800}\selectfont \(\displaystyle 1.0\)}%
\end{pgfscope}%
\begin{pgfscope}%
\definecolor{textcolor}{rgb}{0.317647,0.317647,0.317647}%
\pgfsetstrokecolor{textcolor}%
\pgfsetfillcolor{textcolor}%
\pgftext[x=0.104873in,y=1.586583in,,bottom,rotate=90.000000]{\color{textcolor}\rmfamily\fontsize{6.664000}{7.996800}\selectfont Accuracy}%
\end{pgfscope}%
\begin{pgfscope}%
\pgfpathrectangle{\pgfqpoint{0.345807in}{1.245917in}}{\pgfqpoint{2.378830in}{0.681333in}}%
\pgfusepath{clip}%
\pgfsetbuttcap%
\pgfsetmiterjoin%
\definecolor{currentfill}{rgb}{0.411765,0.411765,0.411765}%
\pgfsetfillcolor{currentfill}%
\pgfsetlinewidth{0.501875pt}%
\definecolor{currentstroke}{rgb}{0.000000,0.000000,0.000000}%
\pgfsetstrokecolor{currentstroke}%
\pgfsetdash{}{0pt}%
\pgfpathmoveto{\pgfqpoint{2.419911in}{1.245917in}}%
\pgfpathlineto{\pgfqpoint{2.518210in}{1.245917in}}%
\pgfpathlineto{\pgfqpoint{2.518210in}{1.909551in}}%
\pgfpathlineto{\pgfqpoint{2.419911in}{1.909551in}}%
\pgfpathclose%
\pgfusepath{stroke,fill}%
\end{pgfscope}%
\begin{pgfscope}%
\pgfpathrectangle{\pgfqpoint{0.345807in}{1.245917in}}{\pgfqpoint{2.378830in}{0.681333in}}%
\pgfusepath{clip}%
\pgfsetbuttcap%
\pgfsetmiterjoin%
\definecolor{currentfill}{rgb}{0.411765,0.411765,0.411765}%
\pgfsetfillcolor{currentfill}%
\pgfsetlinewidth{0.501875pt}%
\definecolor{currentstroke}{rgb}{0.000000,0.000000,0.000000}%
\pgfsetstrokecolor{currentstroke}%
\pgfsetdash{}{0pt}%
\pgfpathmoveto{\pgfqpoint{0.453936in}{1.245917in}}%
\pgfpathlineto{\pgfqpoint{0.552235in}{1.245917in}}%
\pgfpathlineto{\pgfqpoint{0.552235in}{1.318693in}}%
\pgfpathlineto{\pgfqpoint{0.453936in}{1.318693in}}%
\pgfpathclose%
\pgfusepath{stroke,fill}%
\end{pgfscope}%
\begin{pgfscope}%
\pgfpathrectangle{\pgfqpoint{0.345807in}{1.245917in}}{\pgfqpoint{2.378830in}{0.681333in}}%
\pgfusepath{clip}%
\pgfsetbuttcap%
\pgfsetmiterjoin%
\definecolor{currentfill}{rgb}{0.411765,0.411765,0.411765}%
\pgfsetfillcolor{currentfill}%
\pgfsetlinewidth{0.501875pt}%
\definecolor{currentstroke}{rgb}{0.000000,0.000000,0.000000}%
\pgfsetstrokecolor{currentstroke}%
\pgfsetdash{}{0pt}%
\pgfpathmoveto{\pgfqpoint{0.945430in}{1.245917in}}%
\pgfpathlineto{\pgfqpoint{1.043728in}{1.245917in}}%
\pgfpathlineto{\pgfqpoint{1.043728in}{1.296167in}}%
\pgfpathlineto{\pgfqpoint{0.945430in}{1.296167in}}%
\pgfpathclose%
\pgfusepath{stroke,fill}%
\end{pgfscope}%
\begin{pgfscope}%
\pgfpathrectangle{\pgfqpoint{0.345807in}{1.245917in}}{\pgfqpoint{2.378830in}{0.681333in}}%
\pgfusepath{clip}%
\pgfsetbuttcap%
\pgfsetmiterjoin%
\definecolor{currentfill}{rgb}{0.411765,0.411765,0.411765}%
\pgfsetfillcolor{currentfill}%
\pgfsetlinewidth{0.501875pt}%
\definecolor{currentstroke}{rgb}{0.000000,0.000000,0.000000}%
\pgfsetstrokecolor{currentstroke}%
\pgfsetdash{}{0pt}%
\pgfpathmoveto{\pgfqpoint{1.436923in}{1.245917in}}%
\pgfpathlineto{\pgfqpoint{1.535222in}{1.245917in}}%
\pgfpathlineto{\pgfqpoint{1.535222in}{1.294897in}}%
\pgfpathlineto{\pgfqpoint{1.436923in}{1.294897in}}%
\pgfpathclose%
\pgfusepath{stroke,fill}%
\end{pgfscope}%
\begin{pgfscope}%
\pgfpathrectangle{\pgfqpoint{0.345807in}{1.245917in}}{\pgfqpoint{2.378830in}{0.681333in}}%
\pgfusepath{clip}%
\pgfsetbuttcap%
\pgfsetmiterjoin%
\definecolor{currentfill}{rgb}{0.411765,0.411765,0.411765}%
\pgfsetfillcolor{currentfill}%
\pgfsetlinewidth{0.501875pt}%
\definecolor{currentstroke}{rgb}{0.000000,0.000000,0.000000}%
\pgfsetstrokecolor{currentstroke}%
\pgfsetdash{}{0pt}%
\pgfpathmoveto{\pgfqpoint{1.928417in}{1.245917in}}%
\pgfpathlineto{\pgfqpoint{2.026716in}{1.245917in}}%
\pgfpathlineto{\pgfqpoint{2.026716in}{1.313150in}}%
\pgfpathlineto{\pgfqpoint{1.928417in}{1.313150in}}%
\pgfpathclose%
\pgfusepath{stroke,fill}%
\end{pgfscope}%
\begin{pgfscope}%
\pgfpathrectangle{\pgfqpoint{0.345807in}{1.245917in}}{\pgfqpoint{2.378830in}{0.681333in}}%
\pgfusepath{clip}%
\pgfsetbuttcap%
\pgfsetmiterjoin%
\definecolor{currentfill}{rgb}{0.827451,0.827451,0.827451}%
\pgfsetfillcolor{currentfill}%
\pgfsetlinewidth{0.501875pt}%
\definecolor{currentstroke}{rgb}{0.000000,0.000000,0.000000}%
\pgfsetstrokecolor{currentstroke}%
\pgfsetdash{}{0pt}%
\pgfpathmoveto{\pgfqpoint{2.518210in}{1.245917in}}%
\pgfpathlineto{\pgfqpoint{2.616508in}{1.245917in}}%
\pgfpathlineto{\pgfqpoint{2.616508in}{1.895756in}}%
\pgfpathlineto{\pgfqpoint{2.518210in}{1.895756in}}%
\pgfpathclose%
\pgfusepath{stroke,fill}%
\end{pgfscope}%
\begin{pgfscope}%
\pgfpathrectangle{\pgfqpoint{0.345807in}{1.245917in}}{\pgfqpoint{2.378830in}{0.681333in}}%
\pgfusepath{clip}%
\pgfsetbuttcap%
\pgfsetmiterjoin%
\definecolor{currentfill}{rgb}{0.827451,0.827451,0.827451}%
\pgfsetfillcolor{currentfill}%
\pgfsetlinewidth{0.501875pt}%
\definecolor{currentstroke}{rgb}{0.000000,0.000000,0.000000}%
\pgfsetstrokecolor{currentstroke}%
\pgfsetdash{}{0pt}%
\pgfpathmoveto{\pgfqpoint{0.552235in}{1.245917in}}%
\pgfpathlineto{\pgfqpoint{0.650533in}{1.245917in}}%
\pgfpathlineto{\pgfqpoint{0.650533in}{1.297593in}}%
\pgfpathlineto{\pgfqpoint{0.552235in}{1.297593in}}%
\pgfpathclose%
\pgfusepath{stroke,fill}%
\end{pgfscope}%
\begin{pgfscope}%
\pgfpathrectangle{\pgfqpoint{0.345807in}{1.245917in}}{\pgfqpoint{2.378830in}{0.681333in}}%
\pgfusepath{clip}%
\pgfsetbuttcap%
\pgfsetmiterjoin%
\definecolor{currentfill}{rgb}{0.827451,0.827451,0.827451}%
\pgfsetfillcolor{currentfill}%
\pgfsetlinewidth{0.501875pt}%
\definecolor{currentstroke}{rgb}{0.000000,0.000000,0.000000}%
\pgfsetstrokecolor{currentstroke}%
\pgfsetdash{}{0pt}%
\pgfpathmoveto{\pgfqpoint{1.043728in}{1.245917in}}%
\pgfpathlineto{\pgfqpoint{1.142027in}{1.245917in}}%
\pgfpathlineto{\pgfqpoint{1.142027in}{1.295331in}}%
\pgfpathlineto{\pgfqpoint{1.043728in}{1.295331in}}%
\pgfpathclose%
\pgfusepath{stroke,fill}%
\end{pgfscope}%
\begin{pgfscope}%
\pgfpathrectangle{\pgfqpoint{0.345807in}{1.245917in}}{\pgfqpoint{2.378830in}{0.681333in}}%
\pgfusepath{clip}%
\pgfsetbuttcap%
\pgfsetmiterjoin%
\definecolor{currentfill}{rgb}{0.827451,0.827451,0.827451}%
\pgfsetfillcolor{currentfill}%
\pgfsetlinewidth{0.501875pt}%
\definecolor{currentstroke}{rgb}{0.000000,0.000000,0.000000}%
\pgfsetstrokecolor{currentstroke}%
\pgfsetdash{}{0pt}%
\pgfpathmoveto{\pgfqpoint{1.535222in}{1.245917in}}%
\pgfpathlineto{\pgfqpoint{1.633521in}{1.245917in}}%
\pgfpathlineto{\pgfqpoint{1.633521in}{1.294045in}}%
\pgfpathlineto{\pgfqpoint{1.535222in}{1.294045in}}%
\pgfpathclose%
\pgfusepath{stroke,fill}%
\end{pgfscope}%
\begin{pgfscope}%
\pgfpathrectangle{\pgfqpoint{0.345807in}{1.245917in}}{\pgfqpoint{2.378830in}{0.681333in}}%
\pgfusepath{clip}%
\pgfsetbuttcap%
\pgfsetmiterjoin%
\definecolor{currentfill}{rgb}{0.827451,0.827451,0.827451}%
\pgfsetfillcolor{currentfill}%
\pgfsetlinewidth{0.501875pt}%
\definecolor{currentstroke}{rgb}{0.000000,0.000000,0.000000}%
\pgfsetstrokecolor{currentstroke}%
\pgfsetdash{}{0pt}%
\pgfpathmoveto{\pgfqpoint{2.026716in}{1.245917in}}%
\pgfpathlineto{\pgfqpoint{2.125015in}{1.245917in}}%
\pgfpathlineto{\pgfqpoint{2.125015in}{1.302517in}}%
\pgfpathlineto{\pgfqpoint{2.026716in}{1.302517in}}%
\pgfpathclose%
\pgfusepath{stroke,fill}%
\end{pgfscope}%
\begin{pgfscope}%
\pgfpathrectangle{\pgfqpoint{0.345807in}{1.245917in}}{\pgfqpoint{2.378830in}{0.681333in}}%
\pgfusepath{clip}%
\pgfsetbuttcap%
\pgfsetroundjoin%
\pgfsetlinewidth{0.803000pt}%
\definecolor{currentstroke}{rgb}{0.000000,0.000000,0.000000}%
\pgfsetstrokecolor{currentstroke}%
\pgfsetdash{}{0pt}%
\pgfpathmoveto{\pgfqpoint{2.469060in}{1.903089in}}%
\pgfpathlineto{\pgfqpoint{2.469060in}{1.916014in}}%
\pgfusepath{stroke}%
\end{pgfscope}%
\begin{pgfscope}%
\pgfpathrectangle{\pgfqpoint{0.345807in}{1.245917in}}{\pgfqpoint{2.378830in}{0.681333in}}%
\pgfusepath{clip}%
\pgfsetbuttcap%
\pgfsetroundjoin%
\pgfsetlinewidth{0.803000pt}%
\definecolor{currentstroke}{rgb}{0.000000,0.000000,0.000000}%
\pgfsetstrokecolor{currentstroke}%
\pgfsetdash{}{0pt}%
\pgfpathmoveto{\pgfqpoint{0.503085in}{1.317333in}}%
\pgfpathlineto{\pgfqpoint{0.503085in}{1.320054in}}%
\pgfusepath{stroke}%
\end{pgfscope}%
\begin{pgfscope}%
\pgfpathrectangle{\pgfqpoint{0.345807in}{1.245917in}}{\pgfqpoint{2.378830in}{0.681333in}}%
\pgfusepath{clip}%
\pgfsetbuttcap%
\pgfsetroundjoin%
\pgfsetlinewidth{0.803000pt}%
\definecolor{currentstroke}{rgb}{0.000000,0.000000,0.000000}%
\pgfsetstrokecolor{currentstroke}%
\pgfsetdash{}{0pt}%
\pgfpathmoveto{\pgfqpoint{0.994579in}{1.294671in}}%
\pgfpathlineto{\pgfqpoint{0.994579in}{1.297663in}}%
\pgfusepath{stroke}%
\end{pgfscope}%
\begin{pgfscope}%
\pgfpathrectangle{\pgfqpoint{0.345807in}{1.245917in}}{\pgfqpoint{2.378830in}{0.681333in}}%
\pgfusepath{clip}%
\pgfsetbuttcap%
\pgfsetroundjoin%
\pgfsetlinewidth{0.803000pt}%
\definecolor{currentstroke}{rgb}{0.000000,0.000000,0.000000}%
\pgfsetstrokecolor{currentstroke}%
\pgfsetdash{}{0pt}%
\pgfpathmoveto{\pgfqpoint{1.486073in}{1.290586in}}%
\pgfpathlineto{\pgfqpoint{1.486073in}{1.299207in}}%
\pgfusepath{stroke}%
\end{pgfscope}%
\begin{pgfscope}%
\pgfpathrectangle{\pgfqpoint{0.345807in}{1.245917in}}{\pgfqpoint{2.378830in}{0.681333in}}%
\pgfusepath{clip}%
\pgfsetbuttcap%
\pgfsetroundjoin%
\pgfsetlinewidth{0.803000pt}%
\definecolor{currentstroke}{rgb}{0.000000,0.000000,0.000000}%
\pgfsetstrokecolor{currentstroke}%
\pgfsetdash{}{0pt}%
\pgfpathmoveto{\pgfqpoint{1.977567in}{1.298836in}}%
\pgfpathlineto{\pgfqpoint{1.977567in}{1.327463in}}%
\pgfusepath{stroke}%
\end{pgfscope}%
\begin{pgfscope}%
\pgfpathrectangle{\pgfqpoint{0.345807in}{1.245917in}}{\pgfqpoint{2.378830in}{0.681333in}}%
\pgfusepath{clip}%
\pgfsetbuttcap%
\pgfsetroundjoin%
\pgfsetlinewidth{0.803000pt}%
\definecolor{currentstroke}{rgb}{0.000000,0.000000,0.000000}%
\pgfsetstrokecolor{currentstroke}%
\pgfsetdash{}{0pt}%
\pgfpathmoveto{\pgfqpoint{2.567359in}{1.884256in}}%
\pgfpathlineto{\pgfqpoint{2.567359in}{1.907257in}}%
\pgfusepath{stroke}%
\end{pgfscope}%
\begin{pgfscope}%
\pgfpathrectangle{\pgfqpoint{0.345807in}{1.245917in}}{\pgfqpoint{2.378830in}{0.681333in}}%
\pgfusepath{clip}%
\pgfsetbuttcap%
\pgfsetroundjoin%
\pgfsetlinewidth{0.803000pt}%
\definecolor{currentstroke}{rgb}{0.000000,0.000000,0.000000}%
\pgfsetstrokecolor{currentstroke}%
\pgfsetdash{}{0pt}%
\pgfpathmoveto{\pgfqpoint{0.601384in}{1.291295in}}%
\pgfpathlineto{\pgfqpoint{0.601384in}{1.303891in}}%
\pgfusepath{stroke}%
\end{pgfscope}%
\begin{pgfscope}%
\pgfpathrectangle{\pgfqpoint{0.345807in}{1.245917in}}{\pgfqpoint{2.378830in}{0.681333in}}%
\pgfusepath{clip}%
\pgfsetbuttcap%
\pgfsetroundjoin%
\pgfsetlinewidth{0.803000pt}%
\definecolor{currentstroke}{rgb}{0.000000,0.000000,0.000000}%
\pgfsetstrokecolor{currentstroke}%
\pgfsetdash{}{0pt}%
\pgfpathmoveto{\pgfqpoint{1.092878in}{1.288959in}}%
\pgfpathlineto{\pgfqpoint{1.092878in}{1.301703in}}%
\pgfusepath{stroke}%
\end{pgfscope}%
\begin{pgfscope}%
\pgfpathrectangle{\pgfqpoint{0.345807in}{1.245917in}}{\pgfqpoint{2.378830in}{0.681333in}}%
\pgfusepath{clip}%
\pgfsetbuttcap%
\pgfsetroundjoin%
\pgfsetlinewidth{0.803000pt}%
\definecolor{currentstroke}{rgb}{0.000000,0.000000,0.000000}%
\pgfsetstrokecolor{currentstroke}%
\pgfsetdash{}{0pt}%
\pgfpathmoveto{\pgfqpoint{1.584372in}{1.285430in}}%
\pgfpathlineto{\pgfqpoint{1.584372in}{1.302660in}}%
\pgfusepath{stroke}%
\end{pgfscope}%
\begin{pgfscope}%
\pgfpathrectangle{\pgfqpoint{0.345807in}{1.245917in}}{\pgfqpoint{2.378830in}{0.681333in}}%
\pgfusepath{clip}%
\pgfsetbuttcap%
\pgfsetroundjoin%
\pgfsetlinewidth{0.803000pt}%
\definecolor{currentstroke}{rgb}{0.000000,0.000000,0.000000}%
\pgfsetstrokecolor{currentstroke}%
\pgfsetdash{}{0pt}%
\pgfpathmoveto{\pgfqpoint{2.075865in}{1.289967in}}%
\pgfpathlineto{\pgfqpoint{2.075865in}{1.315067in}}%
\pgfusepath{stroke}%
\end{pgfscope}%
\begin{pgfscope}%
\pgfsetrectcap%
\pgfsetmiterjoin%
\pgfsetlinewidth{0.501875pt}%
\definecolor{currentstroke}{rgb}{0.317647,0.317647,0.317647}%
\pgfsetstrokecolor{currentstroke}%
\pgfsetdash{}{0pt}%
\pgfpathmoveto{\pgfqpoint{0.345807in}{1.245917in}}%
\pgfpathlineto{\pgfqpoint{0.345807in}{1.927250in}}%
\pgfusepath{stroke}%
\end{pgfscope}%
\begin{pgfscope}%
\pgfsetrectcap%
\pgfsetmiterjoin%
\pgfsetlinewidth{0.501875pt}%
\definecolor{currentstroke}{rgb}{0.317647,0.317647,0.317647}%
\pgfsetstrokecolor{currentstroke}%
\pgfsetdash{}{0pt}%
\pgfpathmoveto{\pgfqpoint{0.345807in}{1.245917in}}%
\pgfpathlineto{\pgfqpoint{2.724637in}{1.245917in}}%
\pgfusepath{stroke}%
\end{pgfscope}%
\begin{pgfscope}%
\definecolor{textcolor}{rgb}{0.000000,0.000000,0.000000}%
\pgfsetstrokecolor{textcolor}%
\pgfsetfillcolor{textcolor}%
\pgftext[x=1.535222in,y=2.010583in,,base]{\color{textcolor}\rmfamily\fontsize{6.664000}{7.996800}\selectfont (b) Last-time-step loss}%
\end{pgfscope}%
\begin{pgfscope}%
\pgfsetbuttcap%
\pgfsetmiterjoin%
\definecolor{currentfill}{rgb}{0.411765,0.411765,0.411765}%
\pgfsetfillcolor{currentfill}%
\pgfsetlinewidth{0.501875pt}%
\definecolor{currentstroke}{rgb}{0.000000,0.000000,0.000000}%
\pgfsetstrokecolor{currentstroke}%
\pgfsetdash{}{0pt}%
\pgfpathmoveto{\pgfqpoint{0.392074in}{1.840963in}}%
\pgfpathlineto{\pgfqpoint{0.500030in}{1.840963in}}%
\pgfpathlineto{\pgfqpoint{0.500030in}{1.869879in}}%
\pgfpathlineto{\pgfqpoint{0.392074in}{1.869879in}}%
\pgfpathclose%
\pgfusepath{stroke,fill}%
\end{pgfscope}%
\begin{pgfscope}%
\definecolor{textcolor}{rgb}{0.000000,0.000000,0.000000}%
\pgfsetstrokecolor{textcolor}%
\pgfsetfillcolor{textcolor}%
\pgftext[x=0.515452in,y=1.827468in,left,base]{\color{textcolor}\rmfamily\fontsize{5.552000}{6.662400}\selectfont Train}%
\end{pgfscope}%
\begin{pgfscope}%
\pgfsetbuttcap%
\pgfsetmiterjoin%
\definecolor{currentfill}{rgb}{0.827451,0.827451,0.827451}%
\pgfsetfillcolor{currentfill}%
\pgfsetlinewidth{0.501875pt}%
\definecolor{currentstroke}{rgb}{0.000000,0.000000,0.000000}%
\pgfsetstrokecolor{currentstroke}%
\pgfsetdash{}{0pt}%
\pgfpathmoveto{\pgfqpoint{0.392074in}{1.764777in}}%
\pgfpathlineto{\pgfqpoint{0.500030in}{1.764777in}}%
\pgfpathlineto{\pgfqpoint{0.500030in}{1.793694in}}%
\pgfpathlineto{\pgfqpoint{0.392074in}{1.793694in}}%
\pgfpathclose%
\pgfusepath{stroke,fill}%
\end{pgfscope}%
\begin{pgfscope}%
\definecolor{textcolor}{rgb}{0.000000,0.000000,0.000000}%
\pgfsetstrokecolor{textcolor}%
\pgfsetfillcolor{textcolor}%
\pgftext[x=0.515452in,y=1.751283in,left,base]{\color{textcolor}\rmfamily\fontsize{5.552000}{6.662400}\selectfont Valid}%
\end{pgfscope}%
\begin{pgfscope}%
\pgfsetbuttcap%
\pgfsetmiterjoin%
\pgfsetlinewidth{0.000000pt}%
\definecolor{currentstroke}{rgb}{0.000000,0.000000,0.000000}%
\pgfsetstrokecolor{currentstroke}%
\pgfsetstrokeopacity{0.000000}%
\pgfsetdash{}{0pt}%
\pgfpathmoveto{\pgfqpoint{0.345807in}{0.241761in}}%
\pgfpathlineto{\pgfqpoint{2.724637in}{0.241761in}}%
\pgfpathlineto{\pgfqpoint{2.724637in}{0.923094in}}%
\pgfpathlineto{\pgfqpoint{0.345807in}{0.923094in}}%
\pgfpathclose%
\pgfusepath{}%
\end{pgfscope}%
\begin{pgfscope}%
\pgfsetbuttcap%
\pgfsetroundjoin%
\definecolor{currentfill}{rgb}{0.317647,0.317647,0.317647}%
\pgfsetfillcolor{currentfill}%
\pgfsetlinewidth{0.803000pt}%
\definecolor{currentstroke}{rgb}{0.317647,0.317647,0.317647}%
\pgfsetstrokecolor{currentstroke}%
\pgfsetdash{}{0pt}%
\pgfsys@defobject{currentmarker}{\pgfqpoint{0.000000in}{0.000000in}}{\pgfqpoint{0.000000in}{0.013889in}}{%
\pgfpathmoveto{\pgfqpoint{0.000000in}{0.000000in}}%
\pgfpathlineto{\pgfqpoint{0.000000in}{0.013889in}}%
\pgfusepath{stroke,fill}%
}%
\begin{pgfscope}%
\pgfsys@transformshift{0.601384in}{0.241761in}%
\pgfsys@useobject{currentmarker}{}%
\end{pgfscope}%
\end{pgfscope}%
\begin{pgfscope}%
\definecolor{textcolor}{rgb}{0.317647,0.317647,0.317647}%
\pgfsetstrokecolor{textcolor}%
\pgfsetfillcolor{textcolor}%
\pgftext[x=0.429786in,y=0.134935in,left,base]{\color{textcolor}\rmfamily\fontsize{6.664000}{7.996800}\selectfont Spiking}%
\end{pgfscope}%
\begin{pgfscope}%
\definecolor{textcolor}{rgb}{0.317647,0.317647,0.317647}%
\pgfsetstrokecolor{textcolor}%
\pgfsetfillcolor{textcolor}%
\pgftext[x=0.446909in,y=0.038881in,left,base]{\color{textcolor}\rmfamily\fontsize{6.664000}{7.996800}\selectfont 1-layer}%
\end{pgfscope}%
\begin{pgfscope}%
\pgfsetbuttcap%
\pgfsetroundjoin%
\definecolor{currentfill}{rgb}{0.317647,0.317647,0.317647}%
\pgfsetfillcolor{currentfill}%
\pgfsetlinewidth{0.803000pt}%
\definecolor{currentstroke}{rgb}{0.317647,0.317647,0.317647}%
\pgfsetstrokecolor{currentstroke}%
\pgfsetdash{}{0pt}%
\pgfsys@defobject{currentmarker}{\pgfqpoint{0.000000in}{0.000000in}}{\pgfqpoint{0.000000in}{0.013889in}}{%
\pgfpathmoveto{\pgfqpoint{0.000000in}{0.000000in}}%
\pgfpathlineto{\pgfqpoint{0.000000in}{0.013889in}}%
\pgfusepath{stroke,fill}%
}%
\begin{pgfscope}%
\pgfsys@transformshift{1.092878in}{0.241761in}%
\pgfsys@useobject{currentmarker}{}%
\end{pgfscope}%
\end{pgfscope}%
\begin{pgfscope}%
\definecolor{textcolor}{rgb}{0.317647,0.317647,0.317647}%
\pgfsetstrokecolor{textcolor}%
\pgfsetfillcolor{textcolor}%
\pgftext[x=0.921280in,y=0.134935in,left,base]{\color{textcolor}\rmfamily\fontsize{6.664000}{7.996800}\selectfont Spiking}%
\end{pgfscope}%
\begin{pgfscope}%
\definecolor{textcolor}{rgb}{0.317647,0.317647,0.317647}%
\pgfsetstrokecolor{textcolor}%
\pgfsetfillcolor{textcolor}%
\pgftext[x=0.938403in,y=0.038881in,left,base]{\color{textcolor}\rmfamily\fontsize{6.664000}{7.996800}\selectfont 2-layer}%
\end{pgfscope}%
\begin{pgfscope}%
\pgfsetbuttcap%
\pgfsetroundjoin%
\definecolor{currentfill}{rgb}{0.317647,0.317647,0.317647}%
\pgfsetfillcolor{currentfill}%
\pgfsetlinewidth{0.803000pt}%
\definecolor{currentstroke}{rgb}{0.317647,0.317647,0.317647}%
\pgfsetstrokecolor{currentstroke}%
\pgfsetdash{}{0pt}%
\pgfsys@defobject{currentmarker}{\pgfqpoint{0.000000in}{0.000000in}}{\pgfqpoint{0.000000in}{0.013889in}}{%
\pgfpathmoveto{\pgfqpoint{0.000000in}{0.000000in}}%
\pgfpathlineto{\pgfqpoint{0.000000in}{0.013889in}}%
\pgfusepath{stroke,fill}%
}%
\begin{pgfscope}%
\pgfsys@transformshift{1.584372in}{0.241761in}%
\pgfsys@useobject{currentmarker}{}%
\end{pgfscope}%
\end{pgfscope}%
\begin{pgfscope}%
\definecolor{textcolor}{rgb}{0.317647,0.317647,0.317647}%
\pgfsetstrokecolor{textcolor}%
\pgfsetfillcolor{textcolor}%
\pgftext[x=1.412774in,y=0.134935in,left,base]{\color{textcolor}\rmfamily\fontsize{6.664000}{7.996800}\selectfont Spiking}%
\end{pgfscope}%
\begin{pgfscope}%
\definecolor{textcolor}{rgb}{0.317647,0.317647,0.317647}%
\pgfsetstrokecolor{textcolor}%
\pgfsetfillcolor{textcolor}%
\pgftext[x=1.429896in,y=0.038881in,left,base]{\color{textcolor}\rmfamily\fontsize{6.664000}{7.996800}\selectfont 3-layer}%
\end{pgfscope}%
\begin{pgfscope}%
\pgfsetbuttcap%
\pgfsetroundjoin%
\definecolor{currentfill}{rgb}{0.317647,0.317647,0.317647}%
\pgfsetfillcolor{currentfill}%
\pgfsetlinewidth{0.803000pt}%
\definecolor{currentstroke}{rgb}{0.317647,0.317647,0.317647}%
\pgfsetstrokecolor{currentstroke}%
\pgfsetdash{}{0pt}%
\pgfsys@defobject{currentmarker}{\pgfqpoint{0.000000in}{0.000000in}}{\pgfqpoint{0.000000in}{0.013889in}}{%
\pgfpathmoveto{\pgfqpoint{0.000000in}{0.000000in}}%
\pgfpathlineto{\pgfqpoint{0.000000in}{0.013889in}}%
\pgfusepath{stroke,fill}%
}%
\begin{pgfscope}%
\pgfsys@transformshift{2.075865in}{0.241761in}%
\pgfsys@useobject{currentmarker}{}%
\end{pgfscope}%
\end{pgfscope}%
\begin{pgfscope}%
\definecolor{textcolor}{rgb}{0.317647,0.317647,0.317647}%
\pgfsetstrokecolor{textcolor}%
\pgfsetfillcolor{textcolor}%
\pgftext[x=1.904267in,y=0.134009in,left,base]{\color{textcolor}\rmfamily\fontsize{6.664000}{7.996800}\selectfont Spiking}%
\end{pgfscope}%
\begin{pgfscope}%
\definecolor{textcolor}{rgb}{0.317647,0.317647,0.317647}%
\pgfsetstrokecolor{textcolor}%
\pgfsetfillcolor{textcolor}%
\pgftext[x=1.865810in,y=0.037956in,left,base]{\color{textcolor}\rmfamily\fontsize{6.664000}{7.996800}\selectfont recurrent}%
\end{pgfscope}%
\begin{pgfscope}%
\pgfsetbuttcap%
\pgfsetroundjoin%
\definecolor{currentfill}{rgb}{0.317647,0.317647,0.317647}%
\pgfsetfillcolor{currentfill}%
\pgfsetlinewidth{0.803000pt}%
\definecolor{currentstroke}{rgb}{0.317647,0.317647,0.317647}%
\pgfsetstrokecolor{currentstroke}%
\pgfsetdash{}{0pt}%
\pgfsys@defobject{currentmarker}{\pgfqpoint{0.000000in}{0.000000in}}{\pgfqpoint{0.000000in}{0.013889in}}{%
\pgfpathmoveto{\pgfqpoint{0.000000in}{0.000000in}}%
\pgfpathlineto{\pgfqpoint{0.000000in}{0.013889in}}%
\pgfusepath{stroke,fill}%
}%
\begin{pgfscope}%
\pgfsys@transformshift{2.567359in}{0.241761in}%
\pgfsys@useobject{currentmarker}{}%
\end{pgfscope}%
\end{pgfscope}%
\begin{pgfscope}%
\definecolor{textcolor}{rgb}{0.317647,0.317647,0.317647}%
\pgfsetstrokecolor{textcolor}%
\pgfsetfillcolor{textcolor}%
\pgftext[x=2.567359in,y=0.200094in,,top]{\color{textcolor}\rmfamily\fontsize{6.664000}{7.996800}\selectfont LSTM}%
\end{pgfscope}%
\begin{pgfscope}%
\pgfpathrectangle{\pgfqpoint{0.345807in}{0.241761in}}{\pgfqpoint{2.378830in}{0.681333in}}%
\pgfusepath{clip}%
\pgfsetbuttcap%
\pgfsetroundjoin%
\pgfsetlinewidth{0.301125pt}%
\definecolor{currentstroke}{rgb}{0.725490,0.725490,0.725490}%
\pgfsetstrokecolor{currentstroke}%
\pgfsetdash{{0.300000pt}{0.495000pt}}{0.000000pt}%
\pgfpathmoveto{\pgfqpoint{0.345807in}{0.241761in}}%
\pgfpathlineto{\pgfqpoint{2.724637in}{0.241761in}}%
\pgfusepath{stroke}%
\end{pgfscope}%
\begin{pgfscope}%
\pgfsetbuttcap%
\pgfsetroundjoin%
\definecolor{currentfill}{rgb}{0.317647,0.317647,0.317647}%
\pgfsetfillcolor{currentfill}%
\pgfsetlinewidth{0.803000pt}%
\definecolor{currentstroke}{rgb}{0.317647,0.317647,0.317647}%
\pgfsetstrokecolor{currentstroke}%
\pgfsetdash{}{0pt}%
\pgfsys@defobject{currentmarker}{\pgfqpoint{0.000000in}{0.000000in}}{\pgfqpoint{0.013889in}{0.000000in}}{%
\pgfpathmoveto{\pgfqpoint{0.000000in}{0.000000in}}%
\pgfpathlineto{\pgfqpoint{0.013889in}{0.000000in}}%
\pgfusepath{stroke,fill}%
}%
\begin{pgfscope}%
\pgfsys@transformshift{0.345807in}{0.241761in}%
\pgfsys@useobject{currentmarker}{}%
\end{pgfscope}%
\end{pgfscope}%
\begin{pgfscope}%
\definecolor{textcolor}{rgb}{0.317647,0.317647,0.317647}%
\pgfsetstrokecolor{textcolor}%
\pgfsetfillcolor{textcolor}%
\pgftext[x=0.160429in,y=0.209644in,left,base]{\color{textcolor}\rmfamily\fontsize{6.664000}{7.996800}\selectfont \(\displaystyle 0.0\)}%
\end{pgfscope}%
\begin{pgfscope}%
\pgfpathrectangle{\pgfqpoint{0.345807in}{0.241761in}}{\pgfqpoint{2.378830in}{0.681333in}}%
\pgfusepath{clip}%
\pgfsetbuttcap%
\pgfsetroundjoin%
\pgfsetlinewidth{0.301125pt}%
\definecolor{currentstroke}{rgb}{0.725490,0.725490,0.725490}%
\pgfsetstrokecolor{currentstroke}%
\pgfsetdash{{0.300000pt}{0.495000pt}}{0.000000pt}%
\pgfpathmoveto{\pgfqpoint{0.345807in}{0.582428in}}%
\pgfpathlineto{\pgfqpoint{2.724637in}{0.582428in}}%
\pgfusepath{stroke}%
\end{pgfscope}%
\begin{pgfscope}%
\pgfsetbuttcap%
\pgfsetroundjoin%
\definecolor{currentfill}{rgb}{0.317647,0.317647,0.317647}%
\pgfsetfillcolor{currentfill}%
\pgfsetlinewidth{0.803000pt}%
\definecolor{currentstroke}{rgb}{0.317647,0.317647,0.317647}%
\pgfsetstrokecolor{currentstroke}%
\pgfsetdash{}{0pt}%
\pgfsys@defobject{currentmarker}{\pgfqpoint{0.000000in}{0.000000in}}{\pgfqpoint{0.013889in}{0.000000in}}{%
\pgfpathmoveto{\pgfqpoint{0.000000in}{0.000000in}}%
\pgfpathlineto{\pgfqpoint{0.013889in}{0.000000in}}%
\pgfusepath{stroke,fill}%
}%
\begin{pgfscope}%
\pgfsys@transformshift{0.345807in}{0.582428in}%
\pgfsys@useobject{currentmarker}{}%
\end{pgfscope}%
\end{pgfscope}%
\begin{pgfscope}%
\definecolor{textcolor}{rgb}{0.317647,0.317647,0.317647}%
\pgfsetstrokecolor{textcolor}%
\pgfsetfillcolor{textcolor}%
\pgftext[x=0.160429in,y=0.550311in,left,base]{\color{textcolor}\rmfamily\fontsize{6.664000}{7.996800}\selectfont \(\displaystyle 0.5\)}%
\end{pgfscope}%
\begin{pgfscope}%
\pgfpathrectangle{\pgfqpoint{0.345807in}{0.241761in}}{\pgfqpoint{2.378830in}{0.681333in}}%
\pgfusepath{clip}%
\pgfsetbuttcap%
\pgfsetroundjoin%
\pgfsetlinewidth{0.301125pt}%
\definecolor{currentstroke}{rgb}{0.725490,0.725490,0.725490}%
\pgfsetstrokecolor{currentstroke}%
\pgfsetdash{{0.300000pt}{0.495000pt}}{0.000000pt}%
\pgfpathmoveto{\pgfqpoint{0.345807in}{0.923094in}}%
\pgfpathlineto{\pgfqpoint{2.724637in}{0.923094in}}%
\pgfusepath{stroke}%
\end{pgfscope}%
\begin{pgfscope}%
\pgfsetbuttcap%
\pgfsetroundjoin%
\definecolor{currentfill}{rgb}{0.317647,0.317647,0.317647}%
\pgfsetfillcolor{currentfill}%
\pgfsetlinewidth{0.803000pt}%
\definecolor{currentstroke}{rgb}{0.317647,0.317647,0.317647}%
\pgfsetstrokecolor{currentstroke}%
\pgfsetdash{}{0pt}%
\pgfsys@defobject{currentmarker}{\pgfqpoint{0.000000in}{0.000000in}}{\pgfqpoint{0.013889in}{0.000000in}}{%
\pgfpathmoveto{\pgfqpoint{0.000000in}{0.000000in}}%
\pgfpathlineto{\pgfqpoint{0.013889in}{0.000000in}}%
\pgfusepath{stroke,fill}%
}%
\begin{pgfscope}%
\pgfsys@transformshift{0.345807in}{0.923094in}%
\pgfsys@useobject{currentmarker}{}%
\end{pgfscope}%
\end{pgfscope}%
\begin{pgfscope}%
\definecolor{textcolor}{rgb}{0.317647,0.317647,0.317647}%
\pgfsetstrokecolor{textcolor}%
\pgfsetfillcolor{textcolor}%
\pgftext[x=0.160429in,y=0.890977in,left,base]{\color{textcolor}\rmfamily\fontsize{6.664000}{7.996800}\selectfont \(\displaystyle 1.0\)}%
\end{pgfscope}%
\begin{pgfscope}%
\definecolor{textcolor}{rgb}{0.317647,0.317647,0.317647}%
\pgfsetstrokecolor{textcolor}%
\pgfsetfillcolor{textcolor}%
\pgftext[x=0.104873in,y=0.582428in,,bottom,rotate=90.000000]{\color{textcolor}\rmfamily\fontsize{6.664000}{7.996800}\selectfont Accuracy}%
\end{pgfscope}%
\begin{pgfscope}%
\pgfpathrectangle{\pgfqpoint{0.345807in}{0.241761in}}{\pgfqpoint{2.378830in}{0.681333in}}%
\pgfusepath{clip}%
\pgfsetbuttcap%
\pgfsetmiterjoin%
\definecolor{currentfill}{rgb}{0.411765,0.411765,0.411765}%
\pgfsetfillcolor{currentfill}%
\pgfsetlinewidth{0.501875pt}%
\definecolor{currentstroke}{rgb}{0.000000,0.000000,0.000000}%
\pgfsetstrokecolor{currentstroke}%
\pgfsetdash{}{0pt}%
\pgfpathmoveto{\pgfqpoint{2.419911in}{0.241761in}}%
\pgfpathlineto{\pgfqpoint{2.518210in}{0.241761in}}%
\pgfpathlineto{\pgfqpoint{2.518210in}{0.922495in}}%
\pgfpathlineto{\pgfqpoint{2.419911in}{0.922495in}}%
\pgfpathclose%
\pgfusepath{stroke,fill}%
\end{pgfscope}%
\begin{pgfscope}%
\pgfpathrectangle{\pgfqpoint{0.345807in}{0.241761in}}{\pgfqpoint{2.378830in}{0.681333in}}%
\pgfusepath{clip}%
\pgfsetbuttcap%
\pgfsetmiterjoin%
\definecolor{currentfill}{rgb}{0.411765,0.411765,0.411765}%
\pgfsetfillcolor{currentfill}%
\pgfsetlinewidth{0.501875pt}%
\definecolor{currentstroke}{rgb}{0.000000,0.000000,0.000000}%
\pgfsetstrokecolor{currentstroke}%
\pgfsetdash{}{0pt}%
\pgfpathmoveto{\pgfqpoint{0.453936in}{0.241761in}}%
\pgfpathlineto{\pgfqpoint{0.552235in}{0.241761in}}%
\pgfpathlineto{\pgfqpoint{0.552235in}{0.914007in}}%
\pgfpathlineto{\pgfqpoint{0.453936in}{0.914007in}}%
\pgfpathclose%
\pgfusepath{stroke,fill}%
\end{pgfscope}%
\begin{pgfscope}%
\pgfpathrectangle{\pgfqpoint{0.345807in}{0.241761in}}{\pgfqpoint{2.378830in}{0.681333in}}%
\pgfusepath{clip}%
\pgfsetbuttcap%
\pgfsetmiterjoin%
\definecolor{currentfill}{rgb}{0.411765,0.411765,0.411765}%
\pgfsetfillcolor{currentfill}%
\pgfsetlinewidth{0.501875pt}%
\definecolor{currentstroke}{rgb}{0.000000,0.000000,0.000000}%
\pgfsetstrokecolor{currentstroke}%
\pgfsetdash{}{0pt}%
\pgfpathmoveto{\pgfqpoint{0.945430in}{0.241761in}}%
\pgfpathlineto{\pgfqpoint{1.043728in}{0.241761in}}%
\pgfpathlineto{\pgfqpoint{1.043728in}{0.920631in}}%
\pgfpathlineto{\pgfqpoint{0.945430in}{0.920631in}}%
\pgfpathclose%
\pgfusepath{stroke,fill}%
\end{pgfscope}%
\begin{pgfscope}%
\pgfpathrectangle{\pgfqpoint{0.345807in}{0.241761in}}{\pgfqpoint{2.378830in}{0.681333in}}%
\pgfusepath{clip}%
\pgfsetbuttcap%
\pgfsetmiterjoin%
\definecolor{currentfill}{rgb}{0.411765,0.411765,0.411765}%
\pgfsetfillcolor{currentfill}%
\pgfsetlinewidth{0.501875pt}%
\definecolor{currentstroke}{rgb}{0.000000,0.000000,0.000000}%
\pgfsetstrokecolor{currentstroke}%
\pgfsetdash{}{0pt}%
\pgfpathmoveto{\pgfqpoint{1.436923in}{0.241761in}}%
\pgfpathlineto{\pgfqpoint{1.535222in}{0.241761in}}%
\pgfpathlineto{\pgfqpoint{1.535222in}{0.903482in}}%
\pgfpathlineto{\pgfqpoint{1.436923in}{0.903482in}}%
\pgfpathclose%
\pgfusepath{stroke,fill}%
\end{pgfscope}%
\begin{pgfscope}%
\pgfpathrectangle{\pgfqpoint{0.345807in}{0.241761in}}{\pgfqpoint{2.378830in}{0.681333in}}%
\pgfusepath{clip}%
\pgfsetbuttcap%
\pgfsetmiterjoin%
\definecolor{currentfill}{rgb}{0.411765,0.411765,0.411765}%
\pgfsetfillcolor{currentfill}%
\pgfsetlinewidth{0.501875pt}%
\definecolor{currentstroke}{rgb}{0.000000,0.000000,0.000000}%
\pgfsetstrokecolor{currentstroke}%
\pgfsetdash{}{0pt}%
\pgfpathmoveto{\pgfqpoint{1.928417in}{0.241761in}}%
\pgfpathlineto{\pgfqpoint{2.026716in}{0.241761in}}%
\pgfpathlineto{\pgfqpoint{2.026716in}{0.920800in}}%
\pgfpathlineto{\pgfqpoint{1.928417in}{0.920800in}}%
\pgfpathclose%
\pgfusepath{stroke,fill}%
\end{pgfscope}%
\begin{pgfscope}%
\pgfpathrectangle{\pgfqpoint{0.345807in}{0.241761in}}{\pgfqpoint{2.378830in}{0.681333in}}%
\pgfusepath{clip}%
\pgfsetbuttcap%
\pgfsetmiterjoin%
\definecolor{currentfill}{rgb}{0.827451,0.827451,0.827451}%
\pgfsetfillcolor{currentfill}%
\pgfsetlinewidth{0.501875pt}%
\definecolor{currentstroke}{rgb}{0.000000,0.000000,0.000000}%
\pgfsetstrokecolor{currentstroke}%
\pgfsetdash{}{0pt}%
\pgfpathmoveto{\pgfqpoint{2.518210in}{0.241761in}}%
\pgfpathlineto{\pgfqpoint{2.616508in}{0.241761in}}%
\pgfpathlineto{\pgfqpoint{2.616508in}{0.904198in}}%
\pgfpathlineto{\pgfqpoint{2.518210in}{0.904198in}}%
\pgfpathclose%
\pgfusepath{stroke,fill}%
\end{pgfscope}%
\begin{pgfscope}%
\pgfpathrectangle{\pgfqpoint{0.345807in}{0.241761in}}{\pgfqpoint{2.378830in}{0.681333in}}%
\pgfusepath{clip}%
\pgfsetbuttcap%
\pgfsetmiterjoin%
\definecolor{currentfill}{rgb}{0.827451,0.827451,0.827451}%
\pgfsetfillcolor{currentfill}%
\pgfsetlinewidth{0.501875pt}%
\definecolor{currentstroke}{rgb}{0.000000,0.000000,0.000000}%
\pgfsetstrokecolor{currentstroke}%
\pgfsetdash{}{0pt}%
\pgfpathmoveto{\pgfqpoint{0.552235in}{0.241761in}}%
\pgfpathlineto{\pgfqpoint{0.650533in}{0.241761in}}%
\pgfpathlineto{\pgfqpoint{0.650533in}{0.632862in}}%
\pgfpathlineto{\pgfqpoint{0.552235in}{0.632862in}}%
\pgfpathclose%
\pgfusepath{stroke,fill}%
\end{pgfscope}%
\begin{pgfscope}%
\pgfpathrectangle{\pgfqpoint{0.345807in}{0.241761in}}{\pgfqpoint{2.378830in}{0.681333in}}%
\pgfusepath{clip}%
\pgfsetbuttcap%
\pgfsetmiterjoin%
\definecolor{currentfill}{rgb}{0.827451,0.827451,0.827451}%
\pgfsetfillcolor{currentfill}%
\pgfsetlinewidth{0.501875pt}%
\definecolor{currentstroke}{rgb}{0.000000,0.000000,0.000000}%
\pgfsetstrokecolor{currentstroke}%
\pgfsetdash{}{0pt}%
\pgfpathmoveto{\pgfqpoint{1.043728in}{0.241761in}}%
\pgfpathlineto{\pgfqpoint{1.142027in}{0.241761in}}%
\pgfpathlineto{\pgfqpoint{1.142027in}{0.669413in}}%
\pgfpathlineto{\pgfqpoint{1.043728in}{0.669413in}}%
\pgfpathclose%
\pgfusepath{stroke,fill}%
\end{pgfscope}%
\begin{pgfscope}%
\pgfpathrectangle{\pgfqpoint{0.345807in}{0.241761in}}{\pgfqpoint{2.378830in}{0.681333in}}%
\pgfusepath{clip}%
\pgfsetbuttcap%
\pgfsetmiterjoin%
\definecolor{currentfill}{rgb}{0.827451,0.827451,0.827451}%
\pgfsetfillcolor{currentfill}%
\pgfsetlinewidth{0.501875pt}%
\definecolor{currentstroke}{rgb}{0.000000,0.000000,0.000000}%
\pgfsetstrokecolor{currentstroke}%
\pgfsetdash{}{0pt}%
\pgfpathmoveto{\pgfqpoint{1.535222in}{0.241761in}}%
\pgfpathlineto{\pgfqpoint{1.633521in}{0.241761in}}%
\pgfpathlineto{\pgfqpoint{1.633521in}{0.661739in}}%
\pgfpathlineto{\pgfqpoint{1.535222in}{0.661739in}}%
\pgfpathclose%
\pgfusepath{stroke,fill}%
\end{pgfscope}%
\begin{pgfscope}%
\pgfpathrectangle{\pgfqpoint{0.345807in}{0.241761in}}{\pgfqpoint{2.378830in}{0.681333in}}%
\pgfusepath{clip}%
\pgfsetbuttcap%
\pgfsetmiterjoin%
\definecolor{currentfill}{rgb}{0.827451,0.827451,0.827451}%
\pgfsetfillcolor{currentfill}%
\pgfsetlinewidth{0.501875pt}%
\definecolor{currentstroke}{rgb}{0.000000,0.000000,0.000000}%
\pgfsetstrokecolor{currentstroke}%
\pgfsetdash{}{0pt}%
\pgfpathmoveto{\pgfqpoint{2.026716in}{0.241761in}}%
\pgfpathlineto{\pgfqpoint{2.125015in}{0.241761in}}%
\pgfpathlineto{\pgfqpoint{2.125015in}{0.826550in}}%
\pgfpathlineto{\pgfqpoint{2.026716in}{0.826550in}}%
\pgfpathclose%
\pgfusepath{stroke,fill}%
\end{pgfscope}%
\begin{pgfscope}%
\pgfpathrectangle{\pgfqpoint{0.345807in}{0.241761in}}{\pgfqpoint{2.378830in}{0.681333in}}%
\pgfusepath{clip}%
\pgfsetbuttcap%
\pgfsetroundjoin%
\pgfsetlinewidth{0.803000pt}%
\definecolor{currentstroke}{rgb}{0.000000,0.000000,0.000000}%
\pgfsetstrokecolor{currentstroke}%
\pgfsetdash{}{0pt}%
\pgfpathmoveto{\pgfqpoint{2.469060in}{0.922110in}}%
\pgfpathlineto{\pgfqpoint{2.469060in}{0.922880in}}%
\pgfusepath{stroke}%
\end{pgfscope}%
\begin{pgfscope}%
\pgfpathrectangle{\pgfqpoint{0.345807in}{0.241761in}}{\pgfqpoint{2.378830in}{0.681333in}}%
\pgfusepath{clip}%
\pgfsetbuttcap%
\pgfsetroundjoin%
\pgfsetlinewidth{0.803000pt}%
\definecolor{currentstroke}{rgb}{0.000000,0.000000,0.000000}%
\pgfsetstrokecolor{currentstroke}%
\pgfsetdash{}{0pt}%
\pgfpathmoveto{\pgfqpoint{0.503085in}{0.912526in}}%
\pgfpathlineto{\pgfqpoint{0.503085in}{0.915488in}}%
\pgfusepath{stroke}%
\end{pgfscope}%
\begin{pgfscope}%
\pgfpathrectangle{\pgfqpoint{0.345807in}{0.241761in}}{\pgfqpoint{2.378830in}{0.681333in}}%
\pgfusepath{clip}%
\pgfsetbuttcap%
\pgfsetroundjoin%
\pgfsetlinewidth{0.803000pt}%
\definecolor{currentstroke}{rgb}{0.000000,0.000000,0.000000}%
\pgfsetstrokecolor{currentstroke}%
\pgfsetdash{}{0pt}%
\pgfpathmoveto{\pgfqpoint{0.994579in}{0.918868in}}%
\pgfpathlineto{\pgfqpoint{0.994579in}{0.922395in}}%
\pgfusepath{stroke}%
\end{pgfscope}%
\begin{pgfscope}%
\pgfpathrectangle{\pgfqpoint{0.345807in}{0.241761in}}{\pgfqpoint{2.378830in}{0.681333in}}%
\pgfusepath{clip}%
\pgfsetbuttcap%
\pgfsetroundjoin%
\pgfsetlinewidth{0.803000pt}%
\definecolor{currentstroke}{rgb}{0.000000,0.000000,0.000000}%
\pgfsetstrokecolor{currentstroke}%
\pgfsetdash{}{0pt}%
\pgfpathmoveto{\pgfqpoint{1.486073in}{0.898116in}}%
\pgfpathlineto{\pgfqpoint{1.486073in}{0.908849in}}%
\pgfusepath{stroke}%
\end{pgfscope}%
\begin{pgfscope}%
\pgfpathrectangle{\pgfqpoint{0.345807in}{0.241761in}}{\pgfqpoint{2.378830in}{0.681333in}}%
\pgfusepath{clip}%
\pgfsetbuttcap%
\pgfsetroundjoin%
\pgfsetlinewidth{0.803000pt}%
\definecolor{currentstroke}{rgb}{0.000000,0.000000,0.000000}%
\pgfsetstrokecolor{currentstroke}%
\pgfsetdash{}{0pt}%
\pgfpathmoveto{\pgfqpoint{1.977567in}{0.917567in}}%
\pgfpathlineto{\pgfqpoint{1.977567in}{0.924033in}}%
\pgfusepath{stroke}%
\end{pgfscope}%
\begin{pgfscope}%
\pgfpathrectangle{\pgfqpoint{0.345807in}{0.241761in}}{\pgfqpoint{2.378830in}{0.681333in}}%
\pgfusepath{clip}%
\pgfsetbuttcap%
\pgfsetroundjoin%
\pgfsetlinewidth{0.803000pt}%
\definecolor{currentstroke}{rgb}{0.000000,0.000000,0.000000}%
\pgfsetstrokecolor{currentstroke}%
\pgfsetdash{}{0pt}%
\pgfpathmoveto{\pgfqpoint{2.567359in}{0.898128in}}%
\pgfpathlineto{\pgfqpoint{2.567359in}{0.910268in}}%
\pgfusepath{stroke}%
\end{pgfscope}%
\begin{pgfscope}%
\pgfpathrectangle{\pgfqpoint{0.345807in}{0.241761in}}{\pgfqpoint{2.378830in}{0.681333in}}%
\pgfusepath{clip}%
\pgfsetbuttcap%
\pgfsetroundjoin%
\pgfsetlinewidth{0.803000pt}%
\definecolor{currentstroke}{rgb}{0.000000,0.000000,0.000000}%
\pgfsetstrokecolor{currentstroke}%
\pgfsetdash{}{0pt}%
\pgfpathmoveto{\pgfqpoint{0.601384in}{0.619833in}}%
\pgfpathlineto{\pgfqpoint{0.601384in}{0.645892in}}%
\pgfusepath{stroke}%
\end{pgfscope}%
\begin{pgfscope}%
\pgfpathrectangle{\pgfqpoint{0.345807in}{0.241761in}}{\pgfqpoint{2.378830in}{0.681333in}}%
\pgfusepath{clip}%
\pgfsetbuttcap%
\pgfsetroundjoin%
\pgfsetlinewidth{0.803000pt}%
\definecolor{currentstroke}{rgb}{0.000000,0.000000,0.000000}%
\pgfsetstrokecolor{currentstroke}%
\pgfsetdash{}{0pt}%
\pgfpathmoveto{\pgfqpoint{1.092878in}{0.654716in}}%
\pgfpathlineto{\pgfqpoint{1.092878in}{0.684110in}}%
\pgfusepath{stroke}%
\end{pgfscope}%
\begin{pgfscope}%
\pgfpathrectangle{\pgfqpoint{0.345807in}{0.241761in}}{\pgfqpoint{2.378830in}{0.681333in}}%
\pgfusepath{clip}%
\pgfsetbuttcap%
\pgfsetroundjoin%
\pgfsetlinewidth{0.803000pt}%
\definecolor{currentstroke}{rgb}{0.000000,0.000000,0.000000}%
\pgfsetstrokecolor{currentstroke}%
\pgfsetdash{}{0pt}%
\pgfpathmoveto{\pgfqpoint{1.584372in}{0.646402in}}%
\pgfpathlineto{\pgfqpoint{1.584372in}{0.677077in}}%
\pgfusepath{stroke}%
\end{pgfscope}%
\begin{pgfscope}%
\pgfpathrectangle{\pgfqpoint{0.345807in}{0.241761in}}{\pgfqpoint{2.378830in}{0.681333in}}%
\pgfusepath{clip}%
\pgfsetbuttcap%
\pgfsetroundjoin%
\pgfsetlinewidth{0.803000pt}%
\definecolor{currentstroke}{rgb}{0.000000,0.000000,0.000000}%
\pgfsetstrokecolor{currentstroke}%
\pgfsetdash{}{0pt}%
\pgfpathmoveto{\pgfqpoint{2.075865in}{0.817083in}}%
\pgfpathlineto{\pgfqpoint{2.075865in}{0.836017in}}%
\pgfusepath{stroke}%
\end{pgfscope}%
\begin{pgfscope}%
\pgfsetrectcap%
\pgfsetmiterjoin%
\pgfsetlinewidth{0.501875pt}%
\definecolor{currentstroke}{rgb}{0.317647,0.317647,0.317647}%
\pgfsetstrokecolor{currentstroke}%
\pgfsetdash{}{0pt}%
\pgfpathmoveto{\pgfqpoint{0.345807in}{0.241761in}}%
\pgfpathlineto{\pgfqpoint{0.345807in}{0.923094in}}%
\pgfusepath{stroke}%
\end{pgfscope}%
\begin{pgfscope}%
\pgfsetrectcap%
\pgfsetmiterjoin%
\pgfsetlinewidth{0.501875pt}%
\definecolor{currentstroke}{rgb}{0.317647,0.317647,0.317647}%
\pgfsetstrokecolor{currentstroke}%
\pgfsetdash{}{0pt}%
\pgfpathmoveto{\pgfqpoint{0.345807in}{0.241761in}}%
\pgfpathlineto{\pgfqpoint{2.724637in}{0.241761in}}%
\pgfusepath{stroke}%
\end{pgfscope}%
\begin{pgfscope}%
\definecolor{textcolor}{rgb}{0.000000,0.000000,0.000000}%
\pgfsetstrokecolor{textcolor}%
\pgfsetfillcolor{textcolor}%
\pgftext[x=1.535222in,y=1.006428in,,base]{\color{textcolor}\rmfamily\fontsize{6.664000}{7.996800}\selectfont (c) Max-over-time loss}%
\end{pgfscope}%
\end{pgfpicture}%
\makeatother%
\endgroup%

%% file: tab/performance.tex
\begin{tabularx}{\columnwidth}{XXXX}
	\hline
	\multicolumn{2}{c}{\textbf{Architecture}}											& \multicolumn{1}{c}{\textbf{\acrshort{sxxx}}}	& \multicolumn{1}{c}{\textbf{\acrshort{sgoogle}}}	\\
	\hline
	\multicolumn{1}{l}{\multirow{4}{*}{\textbf{SVM}}}	& \multicolumn{1}{l}{Linear}	& \multicolumn{1}{c}{\SI{56.0(04)}{}}			& \multicolumn{1}{c}{\SI{21.6(00)}{}} \\
	\multicolumn{1}{l}{}								& \multicolumn{1}{l}{Poly-2}	& \multicolumn{1}{c}{\SI{48.3(02)}{}}			& \multicolumn{1}{c}{\SI{23.0(00)}{}} \\
	\multicolumn{1}{l}{}								& \multicolumn{1}{l}{Poly-3}	& \multicolumn{1}{c}{\SI{46.7(05)}{}}			& \multicolumn{1}{c}{\SI{23.9(00)}{}} \\
	\multicolumn{1}{l}{}								& \multicolumn{1}{l}{RBF}		& \multicolumn{1}{c}{\SI{60.0(03)}{}}			& \multicolumn{1}{c}{\SI{29.5(00)}{}} \\
	\hline
	\multicolumn{2}{l}{\textbf{LSTM}\tnote{1}}											& \multicolumn{1}{c}{\SI{89.0(02)}{}}			& \multicolumn{1}{c}{\SI{73.0(01)}{}} \\
	\multicolumn{2}{l}{\textbf{CNN}}													& \multicolumn{1}{c}{\SI{92.4(07)}{}}			& \multicolumn{1}{c}{\SI{77.7(02)}{}} \\
	\hline
	\multirow{6}{*}{\textbf{Spiking}\tnote{1}}			& \multicolumn{1}{l}{1-layer}						& \multicolumn{1}{c}{\SI{48.1(16)}{}}			& \multicolumn{1}{c}{\SI{32.5(05)}{}} \\
														& \multicolumn{1}{l}{2-layer}						& \multicolumn{1}{c}{\SI{48.6(09)}{}}			& \multicolumn{1}{c}{\SI{38.5(06)}{}} \\
														& \multicolumn{1}{l}{3-layer}						& \multicolumn{1}{c}{\SI{47.5(23)}{}}			& \multicolumn{1}{c}{\SI{41.0(05)}{}} \\
														& \multicolumn{1}{l}{Recurrent}						& \multicolumn{1}{c}{\SI{71.4(19)}{}}			& \multicolumn{1}{c}{\SI{50.9(11)}{}} \\
														&
														\multicolumn{1}{l}{Recurrent
														(best effort)\tnote{2}}			& \multicolumn{1}{c}{\SI{83.2(13)}{}}			& \\
														& \multicolumn{1}{l}{\acrshort{snu}\tnote{2}} 		& \multicolumn{1}{c}{\SI{79.0(16)}{}}			& \\
	\hline
\end{tabularx}
\begin{tablenotes}\footnotesize
	\item[1] Trained with max-over-time loss
	\item[2] \SI{1024} neurons + data augmentation (optimized time constants and 
		channel number combined with noise injection)
\end{tablenotes}

%% file: discussion.tex
\section{\label{sec:discussion}Discussion}

In this article, we introduced two new public domain spike-based classification datasets to facilitate the quantitative comparison of \glspl{snn}.
Further, we provide a first set of baselines for future comparisons by training a range of spiking and non-spiking classifiers.

With these  developments we address a lack of comprehensive
benchmark datasets for \glspl{snn}.
To advance the neuromorphic computing field, we need a set of benchmarks
that pose real-world challenges to
quantify gains and standardize evaluation across different platforms~\cite{davies2019}.
We view the datasets in this article as our contribution toward this goal.
But, since it is difficult to foresee the pace of future developments, we
acknowledge that the present datasets may not prove final. 
Thus, to facilitate their refinement, extension, and the creation of novel datasets,
we released in addition to the spiking data, our conversion
software\footnotemark[3] and
raw audio datasets\footnotemark[2], 
both under permissive public domain licenses.

Both spiking datasets are based on auditory classification tasks but were derived from data that was acquired in different recording settings.
We chose audio data sets as the basis for our benchmarks because audio has a temporal dimension which makes it a natural choice for spike-based processing.
However, in contrast to movie data, audio requires fewer input channels for a faithful representation, which renders the derived spiking datasets computationally more tractable.

We did not use one of the other existing audio datasets as a basis for the spiking version for different reasons.
For instance, a large body of spoken digits is provided by the TIDIGITS dataset~\cite{leonard1991}.
However, this dataset is only available under a commercial license and we were aiming for fully open datasets.
In contrast, the Free Spoken Digit Dataset~\cite{zohar2018} is available under Creative Commons BY 4.0 license. Since this dataset only contains 2k~recordings with an overall lower recording and alignment quality, we deemed recording \gls{xxx} as a necessary contribution.
Other datasets, such as Mozilla's Common Voice~\cite{mozilla2019}, LibriSpeech~\cite{panayotov2015}, and TED-LIUM~\cite{rousseau2012} are also publicly available.
However, these datasets pose more challenging speech detection problems since they are only aligned at the sentence level. 
Such more challenging tasks are left for future research on functional \glspl{snn}.
The Spoken Wikipedia Corpora~\cite{KHN16}, for instance, also provides alignment
at the word level, but requires further preprocessing such as the dissection of
audio files into separate words.
Moreover, the pure size and imbalance in samples per class render the dataset more challenging.
We therefore left its conversion for future work.

The only existing public domain dataset with word-level alignment, tractable
size, and preprocessing requirements that we were aware of at the time of
writing this manuscript was the \gls{google} dataset.
This is the reason why we chose to base one spiking benchmark on \gls{google} while simultaneously providing the separate and the smaller \gls{xxx} dataset with higher recording quality and alignment precision.
Finally, the high-fidelity recordings of the \gls{xxx} also make it suitable for quantitative evaluation of the impact of noise on network performance, because well-characterized levels of noise can be added.

The spike conversion step consists of a published physical inner-ear model~\cite{sieroka2006} followed by an established hair-cell model~\cite{meddis1988}.
The processing chain is completed by a single layer of \glspl{bc} to increase phase-locking and to decrease the overall number of spikes.
This approach is similar to the publicly available DASDIGIT dataset~\cite{anumula_feature_2018}.
DASDIGIT is composed of recordings from the TIDIGIT dataset~\cite{leonard1991} which have been played to a dynamic audio sensor with $2\times 64$ frequency selective channels.
In contrast to \gls{sxxx} and \gls{sgoogle}, the raw audio files of the TIDIGIT dataset are only available under a commercial license.
Also, the frequency resolution, measured in frequency selective bands of the \gls{bm} model, is about a factor of 10 lower.
As the software used for processing \gls{sxxx} and \gls{sgoogle} datasets is publicly available, it is straight forward to extend the present datasets.
This step is more difficult for DASDIGIT, because it requires a dynamic audio sensor. 

We standardize the conversion step from raw audio signals to spikes by generating spikes from the \gls{xxx} and the \gls{google} audio datasets.
In doing so, we both improve the usability settings and reduce a common source of performance variability due to differences in the preprocessing pipelines of the end-user.

To establish the first set of baselines (\cref{tab:performance}), we trained a range of non-spiking and spiking classifiers on both the \gls{sxxx} as well as the \gls{sgoogle}.
In comparing the performance on the full datasets with performance obtained on reduced spike count datasets, we found that the temporal information available in the spike times can be leveraged for better classification by suitable classifiers.
Moreover, architectures with explicit recurrence, like \glspl{lstm} and \glspl{rsnn}, were the best performing models among all architectures we tested.
Most likely, the reverberating activity through recurrent connections implements the required memory, thereby bridging the gap between neural time constants and audio features.
Therefore, the inclusion of additional state variables evolving on a slower time scale as in~\citet{bellec2018} will be an interesting extension to improve performance of \glspl{snn}.

In the present manuscript, we trained \glspl{snn} using surrogate gradients
in combination with \gls{bptt}~\cite{bohte_error-backpropagation_2011,
bellec2018, shrestha_slayer:_2018, neftci_surrogate_2019, yin_effective_2020, wozniak_deep_2020}. 
However, it is essential to realize that there exists
a plethora of alternative gradient-based approaches based on network translation~\cite{zambrano_efficient_2017, pfeiffer_deep_2018, ruckauer_closing_2019, stockl_classifying_2020},
single spike timing~\cite{mostafa_supervised_2018, goltz_fast_2019},
mean firing rate~\cite{hunsberger_spiking_2015, lee_enabling_2019},
and stochastic approximations~\cite{bengio_estimating_2013, rezende_stochastic_2014,
gardner_supervised_2016, jang_introduction_2019}.
Further, there are biologically inspired online approximations of surrogate gradients~\cite{zenke_superspike:_2018, kaiser_synaptic_2020, bellec_solution_2020} and, 
finally, a body of work has used biologically motivated
\gls{stdp}-like learning rules~\cite{kheradpisheh_stdp-based_2018, 
zhang_plasticity-centric_2018} (see~\citet{tavanaei_deep_2018} for a comprehensive review).
An in-depth comparison of this plethora of approaches was
beyond the scope of this manuscript. 
However, the present datasets might prove useful in facilitating a more detailed
comparison of the work mentioned above. As such, it is left as interesting future work
to study how \gls{stdp} interacts with the present datasets.

Our analysis of \gls{sxxx} and \gls{sgoogle} using \glspl{lstm} and \glspl{snn} showed that the choice of loss functions can have a marked effect on classification performance.
While \glspl{lstm} perform best with a last-time-step loss, in which only the last time step is used to calculate the cross entropy loss, \glspl{snn} achieved their highest accuracy for a max-over-time loss, in which the maximum membrane potential of each readout unit is considered.
A detailed analysis of suitable cost functions for training \glspl{snn} is an interesting direction for future research.

In summary, we have introduced two versatile and open spiking datasets and
conducted a first set of performance measurements using \gls{snn} classifiers.
This constitutes an important step forward toward the more quantitative
comparison of functional \glspl{snn} in-silico both on conventional computers
and neuromorphic hardware.

%% file: appendix.tex
\section{\label{sec:inner-ear}Inner ear model}

\glsreset{bm}
\glsreset{hc}
\glsreset{bc}

Audio data were converted into spikes using a model of the inner ear and the ascending auditory pathway (cf.\ \cref{fig:overview}) which combines a  \gls{bm} model with a population of \glspl{hc} followed by a population of \glspl{bc} for spike generation.
We now describe the model components individually.

\subsection{\label{subsubsec:bm}\Acrlong{bm} model}

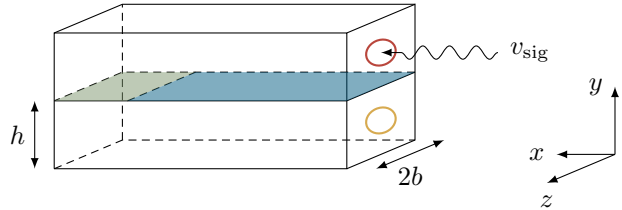
\begin{figure}[t]
	\centering
	\input{graphics/bm}
	\caption{%
		\textbf{Schematic view of the \acrlong{bm} model.}
		The \gls{bm} (blue) separates the \textit{scala tympani} (lower chamber) from the \textit{scala vestibuli} (upper chamber).
		At the \textit{helicotrema} (green), the two scalae are connected.
		The \textit{scala tympani} ends in the \textit{round window} (yellow).
		A sound wave $v_\mathrm{sig}$ is penetrating the eardrum, appling pressure at the \textit{oval window} (red) by moving the \textit{ossicles}, leading to a compression and slower traveling wave.
		We have neglected the \textit{scala media}~\cite{sieroka2006} and consider a stretched form.
	}
	\label{fig:bm}
\end{figure}

As a complete consideration of hydrodynamic \gls{bm} models is beyond the scope
of this manuscript, we closely follow the steps of~\citet{sieroka2006} and highlight the key steps of their derivation.
A fundamental aspect of a cochlea model is the interaction between a fluid and a membrane causing spatial frequency dispersion~\cite{sieroka2006,deBoer1980,deBoer1984}.
Key mechanical features of the cochlea are covered by the simplified geometry of the \gls{bm} in \cref{fig:bm}.
Here, we assumed the fluid to be inviscid and incompressible.
Furthermore, we expect the oscillations to be small that the fluid can be described as linear.
The \gls{bm} was expressed in terms of its mechanical impedance $\xi(x,\omega)$ which depends on the position in the $x$-direction and the angular frequency $\omega = 2\pi\nu$:
\begin{equation}
	\xi(x,\omega) = \frac{1}{i\omega} \left[S(x) -\omega^2 m + i\omega R(x)\right] \, ,
\end{equation}
with a transversal stiffness $S(x) = C_0 e^{-\alpha x} - a$, a resistance $R(x) = R_0 e^{-\alpha x/2}$ and an effective mass $m$~\cite{deBoer1980}.
The damping of the \gls{bm} was described by $\gamma = R_0 / \sqrt{C_0 m}$.
Variations of the stiffness over several orders of magnitude allowed to encompass the entire range of audible frequencies.

Let $p(\boldsymbol{x},\omega)$ be the difference between the pressure in the upper and lower chamber.
The following expression fulfills the boundary conditions $v_y=0$ for $y=h$, and $v_z=0$ at $z=\pm b$~\cite{deBoer1980}:
\begin{multline}
	p(\boldsymbol{x},\omega) = \sum_n \int_0^\infty \frac{\diff k}{2\pi} e^{-ikx} p_0(k) \left[\frac{\cosh(m_0(h-y))}{\cosh(m_0 h)} \right.\\ \cr+ \left. \frac{m_0 \tanh(m_0 h)\cosh(m_1(h-y))}{m_1 \tanh(m_1 h)\cosh(m_1 h)} \cdot\cos\left(\frac{\pi z n}{b}\right)\right] \, . \label{eq:boundary}
\end{multline}
The Laplace equation yields expressions for $m_0=k$ and $m_1=\sqrt{k^2+\pi^2/b^2}$.
Only the principal mode of excitation in the $z$-direction was considered by setting $n=1$.
With the assumptions made above, the Euler equation reads for the $y$-component of the velocity in the middle of the \gls{bm}:
\begin{equation}
	\partial_y p(x,\omega) = -i\omega \rho v_y(x,\omega) = \frac{2i\omega\rho}{\xi(x,\omega)}p(x,\omega) \, , \label{eq:euler}
\end{equation}
where we dropped the $y$ and $z$ argument for readability.
In the following, we consider the limiting case of long waves with $kh \ll 1$.
By combining \cref{eq:boundary,eq:euler}, one gets:
\begin{equation}
	\partial_x^2 p(x,\omega) = \frac{i\omega \rho}{h\xi(x,\omega)} p(x,\omega) \, ,
\end{equation}
where the replacement $\hat{p}(k) \rightarrow p(x)$, $k\rightarrow i\partial_x$ and $k^2\rightarrow \partial_x^2$ has been applied.
Here, $\hat{p}(k)$ denotes the Fourier transform of $p(x,0,0)$.
The solution of this equation was approximated by:
\begin{equation}
  p(x,\omega) = \sqrt{\frac{G(x,\omega)}{g(x,\omega)}} H_0^{(2)}\left(G(x,\omega)\right) \, ,
\end{equation}
where $H_0^{(2)}$ is the second Hankel function and $g(x,\omega)$ and $G(x,\omega)$ are given by:
\begin{align}
	g(x,\omega) &= \omega \sqrt{\frac{\rho}{h\xi(x,\omega)}} \, , \\
	G(x,\omega) &= \int_0^x dx' g(x',\omega)+\frac{2}{\alpha} g(0,\omega) \, .
\end{align}
An analytical expression for $G(x,\omega)$ can be found in~\citet{sieroka2006}.
The model was applied to a given stimulus by:
\begin{equation}
  v_y(x,t) = \int \frac{d\omega}{2\pi} i Z_\mathrm{in} \frac{v_y(x,\omega)}{p(0,\omega)} e^{-i\omega t}v_\mathrm{sig}(\omega) \, ,
\end{equation}
where $v_\mathrm{sig}(\omega)$ is the Fourier transformation of the stimulus.
The input impedance of the cochlea was modeled by:
\begin{equation}
	Z_\mathrm{in}(\omega) = \frac{p(x=0)}{v_x(x=0)} \approx \sqrt{\frac{2C_0}{h}} \frac{i J_0\left(\zeta\right) + Y_0\left(\zeta\right)}{J_1\left(\zeta\right) - iY_1\left(\zeta\right)} \, , \\
\end{equation}
with the Bessel functions of first $J_\beta$ and second $Y_\beta$ kind of order $\beta$ and $\zeta = 2\omega/\alpha \sqrt{2/(hC_0)}$.

To process the audio data, we evaluated $v_y(x,t)$ in the range $(0,\SI{3.5}{\centi\meter}]$ in $N_\mathrm{ch}$ steps of equal size.
Specifically, we chose $N_\mathrm{ch}=700$ as a compromise between a faithful representation of the underlying audio signal and manageable computational cost when using the dataset.
Before applying the \gls{bm} model, each recording $v_\mathrm{sig}(t)$ was normalized to a \gls{rms} value of \SI{0.3}{\centi\meter\per\second}.

\begin{table}[t]
	\centering
	\TableFontSize
	\begin{threeparttable}
		\caption{%
			Model parameters
		}
		\label{tab:params}
		\input{tab/parameter}
	\end{threeparttable}
\end{table}

\subsection{\label{subsubsec:hc}\Acrlong{hc} model}

\begin{figure}[t]
	\centering
	\input{graphics/meddis}
	\caption{%
		\textbf{Schematic view of the transmitter flow within the \glsfirst{hc} model proposed}.
		Figure adapted from~\citet{meddis1986}.
		The model comprises four transmitter pools which allow to describe the transmitter concentration in the synaptic cleft.
	}
	\label{fig:hc}
\end{figure}
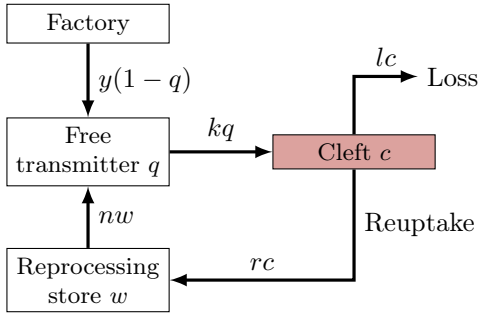

The transformation of the movement of the \gls{bm} to spikes was realized by the \gls{hc}.
The following description illustrates the key steps of~\citet{meddis1986}, to which we refer for further details.

In the \gls{hc} model, one assumes that the cell contains a specific amount of free transmitter molecules $q(x,t)$ which could be released by use of a permeable membrane to the synaptic cleft (\cref{fig:hc}).
The permeability is a function of the velocity of the \gls{bm}, $v_y(x,t)$:
\begin{equation}
  k(x,t) =
  \begin{cases}
    \frac{g\cdot[c\cdot v_y(x,t) + A]}{c\cdot v_y(x,t) + A + B} &\text{for } v_y(x,t) + A > 0 \\
    0 &\text{else}
  \end{cases} \, .
\end{equation}
The amount $c(x,t)$ of transmitter in the cleft is subject to chemical
destruction or loss through diffusion $l\cdot c(x,t)$ as well as re-uptake into the cell $r\cdot c(x,t)$:
\begin{equation}
  \frac{\diff c}{\diff t} = k(x,t) q(x,t) - l\cdot c(x,t) - r\cdot c(x,t) \, .
\end{equation}
A fraction $n\cdot w(x,t)$ of the reuptaken transmitter $w(x,t)$ is continuously transferred to the free transmitter pool:
\begin{equation}
  \frac{\diff w}{\diff t} = r\cdot c(x,t) - n\cdot w(x,t) \, .
\end{equation}
The transmitter originates in a manufacturing base that replenishes the free transmitter pool at a rate $y[1-q(x,t)]$:
\begin{equation}
  \frac{\diff q}{\diff t} = y[1-q(x,t)] + n\cdot w(x,t) - k(x,t) q(x,t) \, 
\end{equation}
While in the cleft, transmitter quanta have a finite probability $\mathcal{P}_\mathrm{spike} = h\cdot c(x,t)\diff t$ of influencing the post-synaptic excitatory potential.
A refractory period was imposed by denying any event which occurs within \SI{1}{\milli\second} of a previous event.
At each position $x$ of the \gls{bm}, we simulated $N_\mathrm{HC} = 40$ independent \glspl{hc}.

\subsection{\label{subsubsec:bc}\Acrlong{bc} model}

The phase-locking of \gls{hc} outputs was increased by feeding their spike output to a population of $N_\mathrm{ch}$ \glspl{bc}.
In contrast to~\cite{rothman1993}, we implemented the \glspl{bc} as standard \gls{lif} neurons.
In more detail, we considered a single layer ($l=1$) of \glspl{bc} without recurrent connections ($V_{ij}^{(l)} = 0$ $\forall i,j$).
The feed-forward weights were set to $W_{ij}^{(l)} = 0.54 / N_\mathrm{HC}\, \forall i,j$.
A single \gls{bc} was used to integrate the spiketrains of the $N_\mathrm{HC} = 40$ \glspl{hc} for each channel of the \gls{bm}.

\section{\label{subsec:controls}Non-spiking classifiers}

For validation purposes, we applied three standard non-spiking methods for time-series classification to the datasets, namely~\glspl{svm}, \glspl{lstm}, and \glspl{cnn}.
We give details on each of them in the following.

\subsection{\label{subsubsec:SVM}\Acrlongpl{svm}}

We trained linear and non-linear \glspl{svm} using \textit{scikit-learn}~\cite{pedregosa2011}.
Specifically, we trained \glspl{svm} with polynomial (up to third degree) and \gls{rbf} kernels.
The vectors in the input space were constructed such that for each sample a $N_\mathrm{ch}$-dimensional vector $\mathbf{x}_i$ is generated by counting the number of spikes emitted by each \gls{bc} in each sample.
Furthermore, features were standardized by removing the mean and scaling to the unit variance.

\subsection{\label{subsubsec:LSTM}Long short term memories}

We used \glspl{lstm} for validation purposes of the temporal data~\cite{hochreiter1997}.
The inputs to the \gls{lstm} consist of the $N_\mathrm{ch}$ spike trains emitted by the \glspl{bc}, but binned in time bins of size \SI{10}{\milli\second}.
We trained \gls{lstm} networks using \textit{TensorFlow 1.14.0} with the \textit{Keras 2.3.0} \gls{api}~\cite{tensorflow2015,chollet2015}.
For all used layers, we stuck to the default parameters and initialization unless mentioned otherwise.
Specifically, we considered a single \gls{lstm} layer with 128 cells with a dropout probability of \SI{0.2}{} for the linear transformation of the input as well as for the linear transformation of the recurrent states.
Last, a readout with softmax activation was applied.
The model was trained with the Adamax optimizer~\cite{kingma2014} and a categorical cross entropy loss defined on the activation of the last time step and the time step with maximal activation.

\subsection{\label{subsubsec:CNN}Convolutional neural networks}

We applied \glspl{cnn} to further test for sperability of the datasets.
To that end, the spike trains were not only binned in time, but also in space.
The temporal binwidth was set to \SI{10}{\milli\second}.
Along the spatial dimension, the data was binned to result in \SI{64}{} distinct input units.
As for \glspl{lstm}, \gls{cnn} networks were trained using \textit{Tensorflow} with the \textit{Keras} \glspl{api} with default parameters and initialization unless mentioned otherwise.
First, a 2D convolution layer with 32 filters of size $ 11\times 11$ and \gls{relu} activation function was applied.
Next, the output was processed by 3 successive blocks, each composed of two 2D convolutional layers, each of them accompanied by batch normalization and \gls{relu} activation.
Both convolutional layers contain 32 filters of size $3\times 3$.
We finalized the blocks by a 2D max-pooling layer with pool size $2\times 2$ and a dropout layer with rate 0.2.
The output of the last of the three blocks was processed by a dense layer with 128 nodes and \gls{relu} activation function followed by a readout with softmax activation.
The whole model is trained with the Adamax optimizer~\cite{kingma2014} and a categorical cross entropy loss for optimization.

%% file: graphics/bm.tex
\begin{tikzpicture}[%
x={(1.1cm,0cm)},
y={(0cm,1cm)},
z={({0.5*cos(25)},{0.5*sin(25)})},
]
\def\a{1.80}
\def\b{3.5}

\coordinate(label) at (3*\b/4,1.5*\a, 0);
\coordinate (A) at (0,0,0); 
\coordinate (B) at (\b,0,0) ;
\coordinate (C) at (\b,\a,0); 
\coordinate (D) at (0,\a,0); 
\coordinate (E) at (0,0,\a); 
\coordinate (F) at (\b,0,\a); 
\coordinate (G) at (\b,\a,\a); 
\coordinate (H) at (0,\a,\a);
\coordinate (I) at (0,\a/2,0);
\coordinate (J) at (\b,\a/2,0);
\coordinate (K) at (\b,\a/2,\a);
\coordinate (L) at (0,\a/2,\a);
\coordinate (M) at (\b/4,\a/2,0);
\coordinate (N) at (\b/4,\a/2,\a);

\coordinate (C0) at (1.8*\b,0,\a/2);
\coordinate (C1) at (1.6*\b,0,\a/2);
\coordinate (C2) at (1.8*\b,0,-\a/2);
\coordinate (C3) at (1.8*\b,\a/2,\a/2);

\coordinate (C4) at (-\b/15,0,0);
\coordinate (C5) at (-\b/15,\a/2,0);
\coordinate (C6) at (\b+\b/10,0,0);
\coordinate (C7) at (\b+\b/10,0,\a);

\coordinate (O1) at (\b,3*\a/4,\a/2);
\coordinate (O2) at (\b,\a/4,\a/2);
\coordinate (O3) at (1.4*\b,3*\a/4,\a/2);

\node[left=1pt of C1]{$x$};
\node[below=1pt of C2]{$z$};
\node[left=1pt of C3]{$y$};

\draw[] (A) -- (B) -- (C) -- (D) -- (A);
\draw[] (I) -- (J) -- (K);
\draw[] (B) -- (F) -- (G) -- (C);
\draw[] (G) -- (H) -- (D);
\draw[densely dashed] (A) -- (E) -- (F);
\draw[densely dashed] (E) -- (H);
\draw[densely dashed] (K) -- (L) -- (I);
\draw[densely dashed] (M) -- (N);
\draw[-latex] (C0) -- (C1);
\draw[-latex] (C0) -- (C2);
\draw[-latex] (C0) -- (C3);
\draw[latex-latex] (C4) -- node[left] {$h$}(C5);
\draw[latex-latex] (C6) -- node[below=1pt] {$2b$}(C7);

\path[fill=green, opacity=0.5] (I) -- (M) -- (N) -- (L) -- (I);
\path[fill=blue, opacity=0.5] (M) -- (J) -- (K) -- (N) -- (M);

\node[ellipse, minimum width=0.4cm, rotate=25, draw=red, thick] (O) at (O1) {};
\node[ellipse, minimum width=0.4cm, rotate=25, draw=orange, thick] (P) at (O2) {};

\draw[-latex,decorate,decoration={snake,post length=1mm}] (O3) -- (O1);
\node[right=1pt of O3]{$v_\mathrm{sig}$};

\end{tikzpicture}

%% file: tab/parameter.tex
\begin{tabular}{llll}
	\hline
	\textbf{Parameter}		& \textbf{Symbol}		& \multicolumn{2}{c}{\textbf{Value}} \\
	\hline
	Damping const.			& $\gamma$				& \multicolumn{2}{l}{\SI{0.15}{\per\second}} \\
	Greenwoods const.		& $a$					& \multicolumn{2}{l}{\SI{35}{\kilo\gram\per\square\second\per\square\centi\meter}} \\
	Stiffness const.		& $C_0$					& \multicolumn{2}{l}{\SI{e9}{\gram\per\square\second\per\square\centi\meter}} \\
	Fluid density			& $\rho$				& \multicolumn{2}{l}{\SI{1.0}{\gram\per\cubic\centi\meter}} \\
	Attenuation factor		& $\alpha$ 				& \multicolumn{2}{l}{\SI{3.0}{\per\centi\meter}} \\
	Height of scala			& $h$					& \multicolumn{2}{l}{\SI{0.1}{\centi\meter}} \\
	Effective mass			& $m$					& \multicolumn{2}{l}{\SI{0.05}{\gram\per\square\centi\meter}} \\
	Number of channels		& $N_\mathrm{ch}$			& \multicolumn{2}{l}{\SI{700}{}} \\
	\hline
	Input scaling factor	& $c$					& \multicolumn{2}{l}{\SI{1.0}{\second\per\centi\meter}} \\
	Permeability offset		& $A$					& \multicolumn{2}{l}{\SI{5}{}} \\
	Permeability rate		& $B$					& \multicolumn{2}{l}{\SI{300}{}} \\
	Maximum permeability	& $g$					& \multicolumn{2}{l}{\SI{1000}{}} \\
	Replenishing rate		& $y$					& \multicolumn{2}{l}{\SI{11.11}{}} \\
	Loss rate				& $l$					& \multicolumn{2}{l}{\SI{1250}{}} \\
	Reuptake rate			& $r$					& \multicolumn{2}{l}{\SI{16667}{}} \\
	Reprocessing rate		& $n$					& \multicolumn{2}{l}{\SI{250}{}} \\
	Propability scaling		& $h$					& \multicolumn{2}{l}{\SI{50000}{}} \\
	Number of \Acrshortpl{hc} at same position & $N_\mathrm{HC}$		& \multicolumn{2}{l}{\SI{40}{}} \\
	\hline
	Synaptic time const.	& $\tau_\mathrm{syn}$	& \SI{0.5}{\milli\second}\tnote{1}	& \SI{10}{\milli\second}\tnote{2} \\
	Membrane time const.	& $\tau_\mathrm{mem}$	& \SI{1}{\milli\second}\tnote{1}	& \SI{20}{\milli\second}\tnote{2} \\
	Refractory time	const.	& $\tau_\mathrm{ref}$	& \SI{1}{\milli\second}\tnote{1}	& \SI{0}{\milli\second}\tnote{2} \\
	Leak potential			& $u_\mathrm{leak}$		& \multicolumn{2}{l}{\SI{0}{}} \\
	Reset potential			& $u_\mathrm{reset}$	& \multicolumn{2}{l}{\SI{0}{}} \\
	Threshold potential		& $u_\mathrm{thres}$	& \multicolumn{2}{l}{\SI{1}{}} \\
	Number of neurons per layer	& $N$			& \multicolumn{2}{l}{\SI{128}{}} \\
	\hline
	Simulation step size				& $\delta t$	& \multicolumn{2}{l}{\SI{0.5}{\milli\second}} \\
	Simulation duration					& $T$			& \multicolumn{2}{l}{\SI{1.0}{\second}} \\
	Batch size					& $N_\mathrm{batch}$		& \multicolumn{2}{l}{\SI{256}{}} \\
	\hline
	Learning rate						& $\eta$		& \multicolumn{2}{l}{\SI{0.001}{}} \\
	Steepness of gradient				& $\beta$		& \multicolumn{2}{l}{\SI{100}{}} \\
	Regularization lower threshold		& $\theta_l$	& \multicolumn{2}{l}{\SI{0.01}{}} \\
	Regularization lower strength		& $s_l$			& \multicolumn{2}{l}{\SI{1.0}{}} \\
	Regularization upper threshold		& $\theta_u$	& \multicolumn{2}{l}{\SI{100.0}{}} \\
	Regularization upper strength		& $s_u$			& \multicolumn{2}{l}{\SI{0.06}{}} \\
	First moment estimates decay rate	& $\beta_1$		& \multicolumn{2}{l}{\SI{0.9}{}} \\
	Second moment estimates decay rate	& $\beta_2$		& \multicolumn{2}{l}{\SI{0.999}{}} \\
	\hline
\end{tabular}
\begin{tablenotes}\footnotesize
	\item[1] \Acrshort{bc} parameter
	\item[2] \Acrshort{snn} parameter
\end{tablenotes}

%% file: graphics/meddis.tex
\begin{tikzpicture}[
		med/.style = {rectangle, text centered, text width=1.9cm, draw=black, font=\small}
	]

    \node (FA) [med] {Factory};
    \node (FT) [med, below of=FA, yshift=-0.7cm] {Free transmitter $q$};
    \node (C)  [med, right of=FT, xshift=2.5cm, fill=red!50] {Cleft $c$};
    \node (L)  [right of=C, xshift=0.3cm,yshift=1.0cm] {Loss};
    \node (RS) [med, below of=FT, yshift=-0.7cm] {Re\-processing store $w$};

    \draw [very thick,-latex] (FA) -- (FT) node [midway,right] {$y(1-q)$};
    \draw [very thick,-latex] (FT) -- (C)  node [midway,above] {$kq$};
    \draw [very thick,-latex] (C)  |- (L)  node [near end,above] {$lc$};
    \draw [very thick,-latex] (C)  |- (RS) node [near end,above] {$rc$} node [near start, right] {Reuptake};
    \draw [very thick,-latex] (RS) -- (FT) node [midway,right] {$nw$};

\end{tikzpicture}

%% file: acknowledgments.tex
\section*{Acknowledgments}

This work has received funding from the European Union Sixth Framework Programme ([FP6/2002-2006]) under grant agreement no 15879 (FACETS),
the European Union Seventh Framework Programme ([FP7/2007-2013]) under grant agreement no 604102 (HBP), 269921 (BrainScaleS) and 243914 (Brain-i-Nets)
and the Horizon 2020 Framework Programme ([H2020/2014-2020]) under grant agreement no 720270 and 785907 (HBP) as well as the Manfred St\"{a}rk Foundation.
The authors acknowledge support by the state of Baden-W\"{u}rttemberg through bwHPC and the German Research Foundation (DFG) through grant no INST 39/963-1 FUGG (bwForCluster NEMO).
This work was supported by the Novartis Research Foundation.
Prof.\ Dr.\ Sebastian Hoth is acknowledged for enabling access to the sound-shielded room at the university hospital Heidelberg.
Prof.\ Dr.\ Hans G\"unter Dosch is acknowledged for discussions and advice on auditory preprocessing. 
David Schumann is acknowledged for mastering the audio files of \gls{xxx}.